\documentclass[twocolumn]{svjour3}         
\smartqed  
\usepackage{graphicx}
\usepackage{amsmath,amsfonts,amssymb}
\usepackage{subfigure}
\usepackage{natbib}
\usepackage{algorithm,algorithmic}
\usepackage{lineno}
\usepackage{color}
\usepackage{url}

\newcommand{\R}{\mathbb{R}}
\newcommand{\G}{\mathbb{G}}

\newcommand{\dt}{\Delta\tau}

\renewcommand{\Re}{\mathbb{R}}
\renewcommand{\vec}{\mathrm{vec}}

\newcommand{\sign}{\mathrm{sign}}

\renewcommand{\mathbf}{\boldsymbol}

\journalname{International Journal of Computer Vision}
\begin{document}

\title{TILT: Transform Invariant Low-rank Textures
}


\author{Zhengdong Zhang \and Arvind Ganesh \and  Xiao Liang \and Yi Ma
}


\institute{Zhengdong Zhang \at Department of Computer Science and Technology, Tsinghua University, Beijing, PRC \\ \email{zhangzdfaint@gmail.com} \and 
Arvind Ganesh (corresponding author) \at
             Department of Electrical and Computer Engineering, University of Illinois at Urbana-Champaign, Urbana, IL, USA \\
              \email{abalasu2@illinois.edu}   \and
              Xiao Liang \at Institute for Advanced Study, Tsinghua University, Beijing, PRC \\
              \email{liangx04@mails.tsinghua.edu.cn}      \and
              Yi Ma \at Department of Electrical and Computer Engineering, University of Illinois at Urbana-Champaign, Urbana, IL, USA \\
              Visual Computing Group, Microsoft Research Asia, Beijing, PRC
              \email{yima@illinois.edu}
}

\date{Received: date / Accepted: date}

\maketitle

\begin{abstract}
In this paper, we show how to efficiently and effectively extract a class of
``low-rank textures'' in a 3D scene from 2D images despite significant corruptions and warping. The low-rank textures capture geometrically meaningful structures in an image, which encompass conventional local features such as edges and corners as well as all kinds  of regular, symmetric patterns ubiquitous in urban environments and man-made objects. Our approach to finding these low-rank textures leverages the recent breakthroughs in convex optimization that enable robust recovery of a high-dimensional low-rank matrix despite gross sparse errors. In the case of planar regions
with significant affine or projective deformation, our method can accurately recover both the
intrinsic low-rank texture and the precise domain transformation, and hence the 3D geometry and
appearance of the planar regions. Extensive experimental
results demonstrate that this new technique works effectively for many regular and near-regular
patterns or objects that are approximately low-rank, such as symmetrical patterns, building facades, printed texts, and human faces.
\end{abstract}

\section{Introduction}
\label{intro}
One of the fundamental problems in computer vision is to identify certain feature points or salient regions in images. These points and regions are the basic building blocks for almost all high-level vision applications such as image matching, 3D reconstruction, object recognition, and scene understanding. Through the years, a large number of methods have been proposed in literature for extracting various types of feature points or salient regions. The detected points or regions typically represent parts of the image that have distinctive geometric or statistical properties such as Canny edges \citep{Canny1986-PAMI}, Harris corners \citep{Harris1988-Alvey}, and textons \citep{Leung2001-IJCV}.

One of the important applications of detecting feature points or regions in images is to establish point-wise correspondences or measure similarity between different images of the same object. This problem is especially challenging if the images are taken from different viewpoints under different lighting conditions. Thus, it is desirable that the detected points/regions are somewhat stable or invariant under transformations incurred by changes in viewpoint or illumination. In the past two decades, numerous ``invariant'' features and descriptors have been proposed, studied, compared, and combined in the literature  (see \citep{Schmid-PAMI05,Winder2007-CVPR} and references therein). Some of the earliest work in this genre were based on using a Markov model to study dependences between various wavelet subbands for rotation invariant textures \citep{Cohen1991-PAMI,Chen1994-PAMI,Wu1996-TIP,Do2002-Multimedia}. There has also been a lot of study in using different kinds of basis functions, such as Gabor wavelets, to filter the image and compute rotation invariant features from the filtered image (see \citep{Haley1999-TIP,Greenspan1994-ICPR,Madiraju1994-ICIP} and references therein).

A widely used invariant feature descriptor is the {\em scale invariant feature transform} (SIFT) \citep{SIFT}, which to a large extent is invariant to changes in rotation and scale ({\it i.e.}, similarity transformations) and illumination. Nevertheless, if the images are shot from very different viewpoints, SIFT is not very successful in establishing reliable correspondences. This problem has been partially addressed by its affine-invariant version \citep{Affine-Scale-SIFT,Affine-SIFT}. However, even these extensions of SIFT are limited in practice: Although the deformation of a small distant patch can be well-approximated by an affine transform, projective transformations are necessary to describe the deformation of a large region viewed through a perspective camera. There has been relatively limited work on projection invariant texture representation \citep{Chang1987-PRL,Kondepudy1994-ICIP}. To the best of our knowledge, from a practical standpoint, there are no feature descriptors that are truly invariant (or even approximately so) under projective transformations or homographies. In addition, these methods normally do not deal with other concurrent nuisance factors such as illumination changes or partial occlusions and corruptions that could severely undermine feature extraction from real images.

\begin{figure*}[!ht]
\centerline{
\subfigure[Input ($r = 35$)]{
\includegraphics[width=0.23\textwidth]{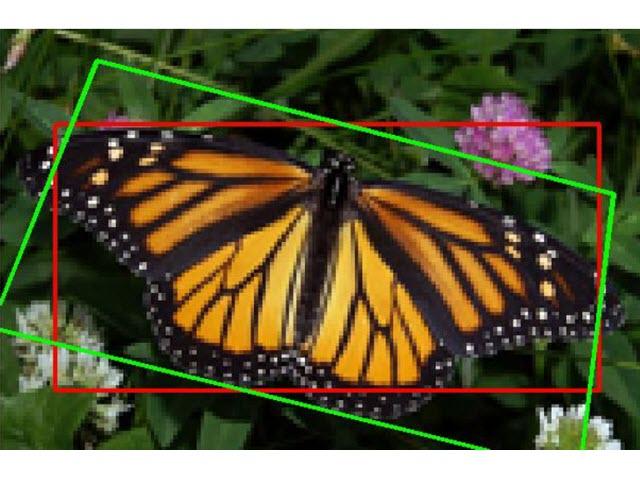}
}
\subfigure[Input ($r = 15$)]{
\includegraphics[width=0.23\textwidth]{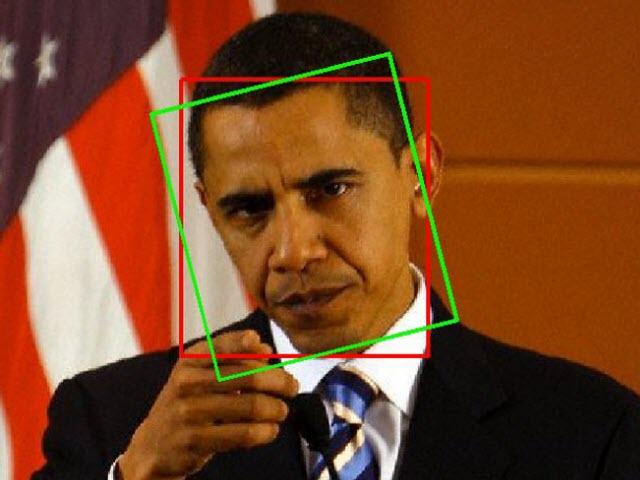}
}
\subfigure[Input ($r = 53$)]{
\includegraphics[width=0.23\textwidth]{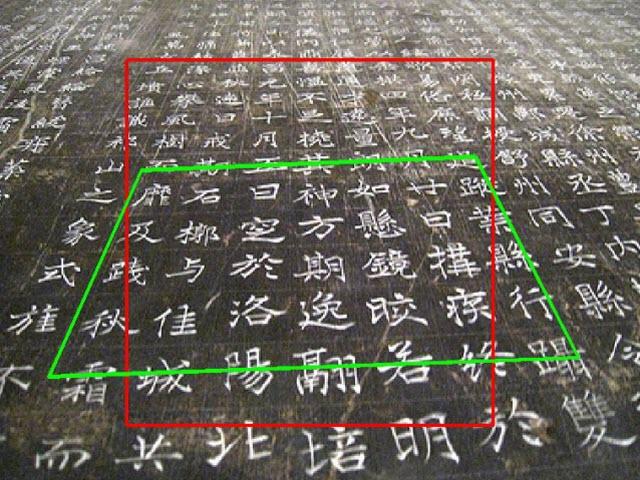}
}
\subfigure[Input ($r = 13$)]{
\includegraphics[width=0.23\textwidth]{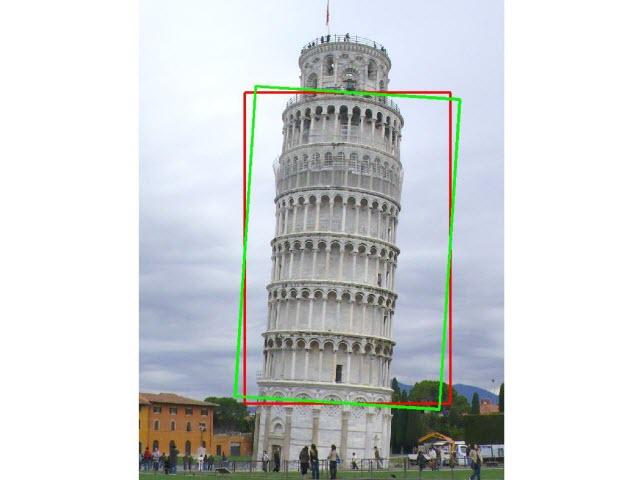}
}
}

\centerline{
\subfigure[Output ($r = 14$)]{
\includegraphics[width=0.23\textwidth]{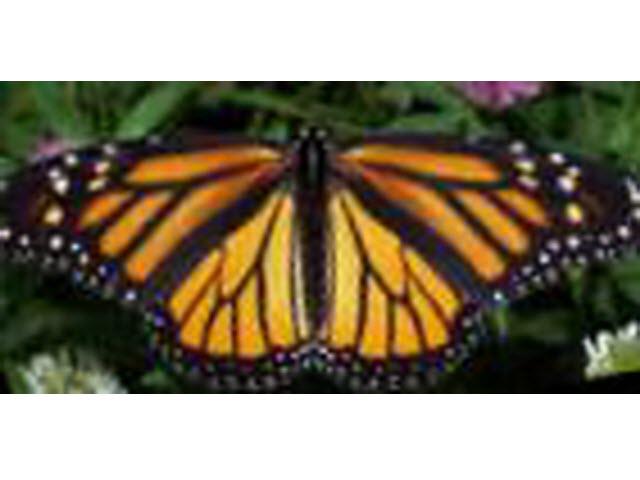}
}
\subfigure[Output ($r = 8$)]{
\includegraphics[width=0.23\textwidth]{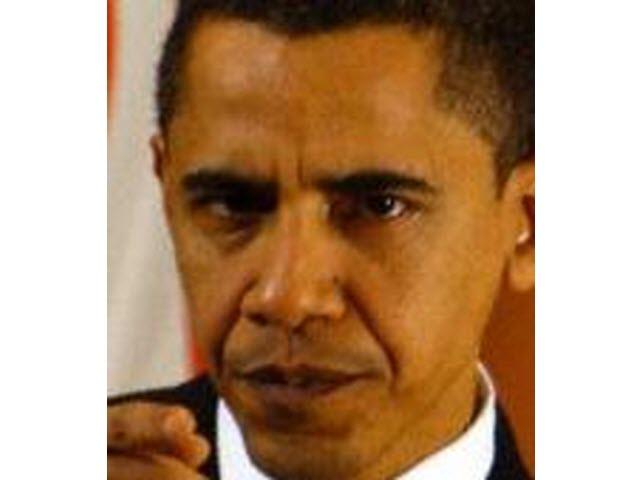}
}
\subfigure[Output ($r = 19$)]{
\includegraphics[width=0.23\textwidth]{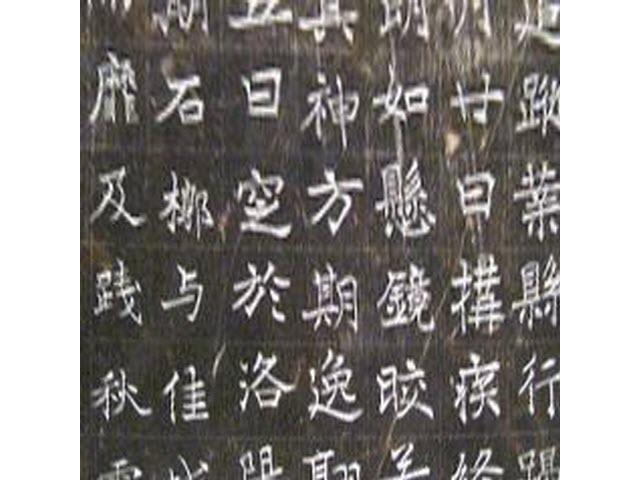}
}
\subfigure[Output ($r = 6$)]{
\includegraphics[width=0.23\textwidth]{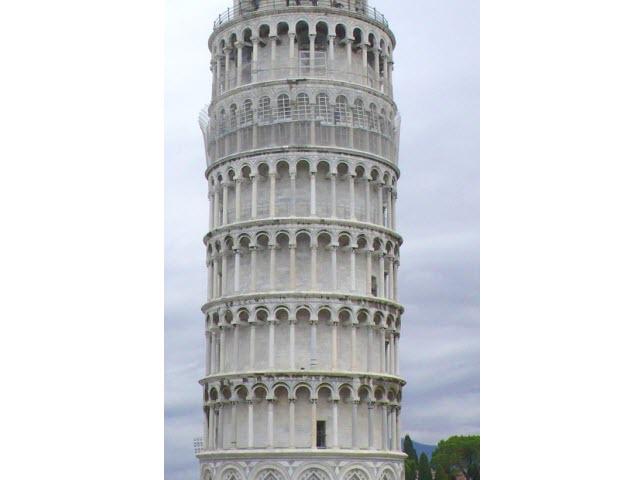}
}
}
\caption{{\bf Low-rank Textures Automatically Rectified by Our Method.} From left to right: a butterfly; a face; a tablet of Chinese characters; and the Leaning Tower of Pisa. Top: red windows denote the original input, green windows denote the deformed texture returned by our method; Bottom: textures in the green window rectified for display. We notice that the rank of the image matrix, denoted by $r$, is much lower for the rectified textures.\label{fig:TILT-overview}}
\end{figure*}

Despite tremendous effort in the past few decades to search for better and richer classes of invariant features in images, there seems to be a fundamental dilemma that none of the existing methods have been able to resolve: On the one hand, if we consider typical classes of transformations incurred on the image domain by changing camera viewpoint and on the image intensity by changing contrast or illumination, then in a strict mathematical sense, {\em invariants of the 2D image are extremely sparse and scarce} -- essentially only the topology of the extrema of the image function remains invariant, known as {\em attributed Reeb tree} (ART) \citep{Soatto-CVPR09}. The numerous ``invariant'' image features proposed in the computer vision literature, including the ones mentioned above, are at best approximately invariant, and often only to a limited extent. On the other hand, {\em a typical 3D scene is rich in regular structures that are full of invariants} (with respect to 3D Euclidean transformations or other well-behaved deformation groups). For instance, in an urban environment, the scene is typically filled with man-made objects that have parallel edges, right-angled corners, regular shapes, symmetric structures, and repeated patterns (see Figures \ref{fig:TILT-overview} and \ref{fig:TILT-examples}). These geometric structures are rich in properties that are invariant under all types of subgroups of the 3D Euclidean group. As a result, their 2D (affine or perspective) images encode very rich and precise information about the 3D geometry and structure of the objects in the scene \citep{Ma-2004,Kosecka-2005,Liu-CVPR08}.


In this paper we propose a technique that aims to resolve the above dilemma about invariant features. We contend that instead of trying to seek local invariant features of the image that are either scarce or imprecise, we should 
\begin{quote}
{\em aim to directly extract certain invariant structures in 3D through their 2D images by undoing the (affine or projective) domain transformations.}
\end{quote}
That is, we cast our quest for ``invariance'' directly as an inverse problem of recovering 3D information from 2D images. However, to solve such challenging inverse problems, we will need some new powerful computational tools which we will introduce and develop in this paper.

Many methods have been developed in the past to detect and extract all types of regular, symmetric patterns from images under affine or projective transforms (see \citep{Liu-CVPR08b} for a recent evaluation). As symmetry is not a property that depends on a small neighborhood of a pixel, it can only be detected from a relatively large region of the image. However, almost all existing methods for detecting symmetric regions and patterns start by extracting and putting together local features such as  corners and edges \citep{Yang-2005} or more advanced local features such as SIFT points \citep{Liu-CVPR08}. As feature detection and edge extraction themselves are sensitive to local image variations such as noise, occlusion, and illumination change, such symmetry detection methods inherently lack robustness and stability. In addition, as we will see in this paper, many regular structures and symmetric patterns do not even have distinctive features. Thus, we need a more general, effective, and robust way of detecting and extracting regular structures in images despite significant distortion and corruption.

Our goal in this paper is to extract invariant information from regions in a 2D image that correspond to a very rich class of regular patterns on a planar surface in 3D, whose appearance can be modeled (approximately) as a ``low-rank'' matrix (see Figure \ref{fig:TILT-overview} for some examples). In some sense, many conventional features mentioned above such as edges, corners, symmetric patterns can all be considered as special instances of such low-rank textures (see Figure \ref{fig:TILT-examples}). Clearly, an image of such a texture may be deformed by the camera projection and undergoes certain domain transformation (say affine or projective). The transformed texture, viewed as a matrix, in general is no longer low-rank in the image domain. Nevertheless, by utilizing advanced convex optimization tools from matrix rank minimization, we will show how to simultaneously recover such a low-rank texture from its deformed image and the associated deformation.

Our method directly uses raw pixel values of the image (window) and there is no need for any pre-extraction of low-level, local features such as corners, edges, SIFT, and DoG features. The proposed solution and algorithm are inherently robust to gross errors caused by corruption, occlusion, or cluttered background as long as they affect a small fraction of the image pixels. Furthermore, our method applies to any image region where there are sufficient low-rank textures, regardless of the size of their spatial support. Thus, we are able to rectify not only small local patches around an edge or a corner but also large global symmetric regions such as an entire facade of a building. We believe that this is a very powerful new tool that allows people to accurately extract rich structural and geometric information about the 3D scene from its 2D images, that are truly invariant of image domain transformations.

\paragraph{Organization of this paper:} The remainder of this paper is organized as follows:  Section \ref{sec:formulation} gives a rigorous definition of ``low-rank textures'' as well as formulates the mathematical problem associated with extracting such textures. Section \ref{sec:solution} gives an efficient and effective algorithm for solving the problem. We provide extensive experimental results to verify the efficacy of the proposed algorithm as well as the usefulness of the extracted low-rank textures in Section 4. In Section 5, we discuss some potential extensions and variations to the basic formulation.

%

\section{Transform Invariant Low-rank Textures}\label{sec:formulation}

\subsection{Definition of Low-rank Textures}
\begin{figure*}[!ht]
\centerline{
\subfigure[Input ($r = 11$)]{
\includegraphics[width=0.22\textwidth]{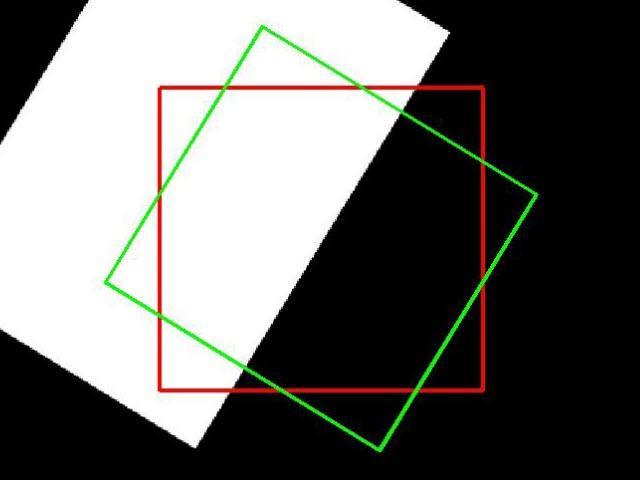}
}
\subfigure[Input ($r = 16$)]{
\includegraphics[width=0.22\textwidth]{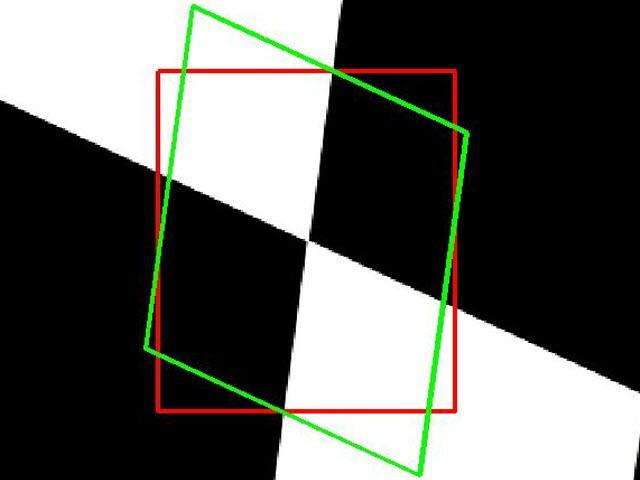}
}
\subfigure[Input ($r = 10$)]{
\includegraphics[width=0.22\textwidth]{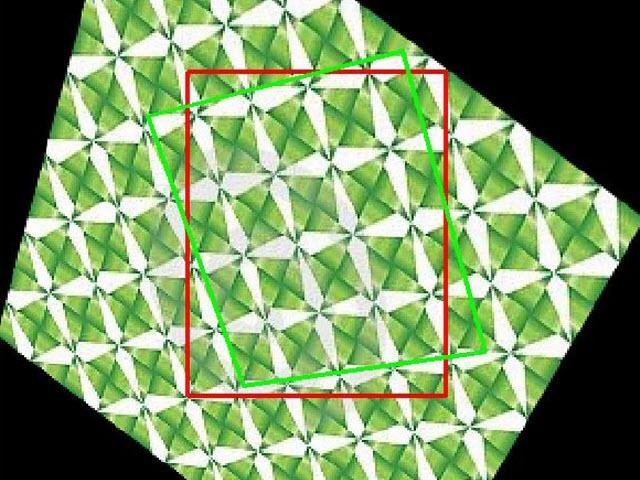}
}
\subfigure[Input ($r = 24$)]{
\includegraphics[width=0.22\textwidth]{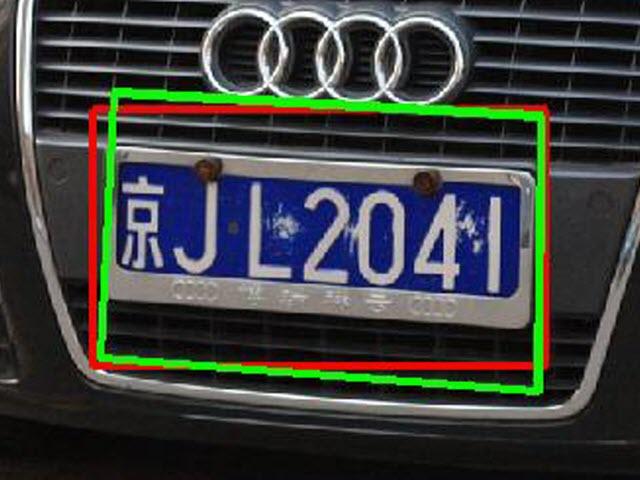}
}
}

\centerline{
\subfigure[Output ($r = 1$)]{
\includegraphics[width=0.22\textwidth]{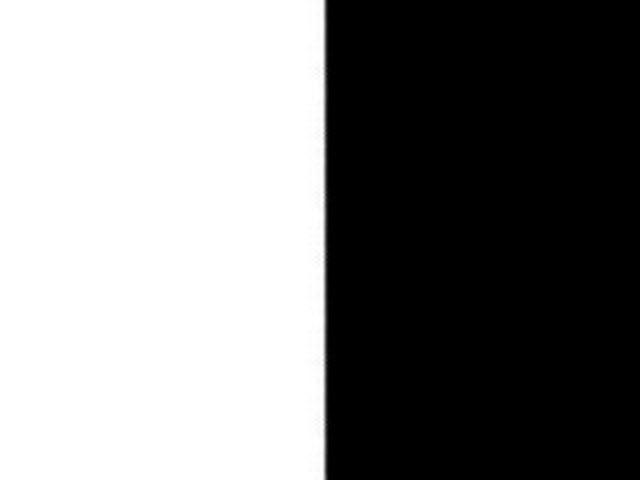}
}
\subfigure[Output ($r = 2$)]{
\includegraphics[width=0.22\textwidth]{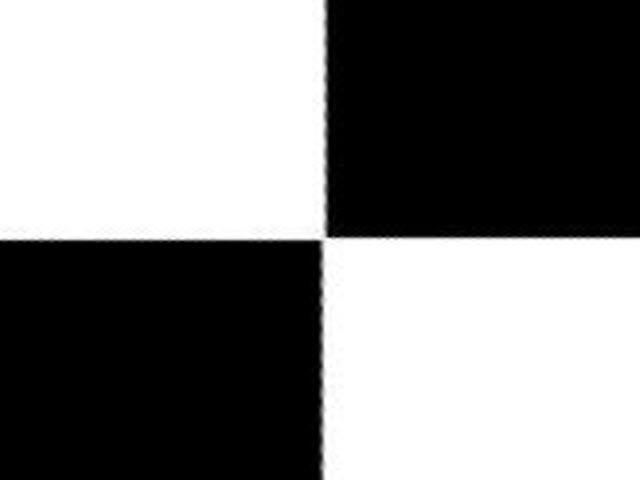}
}
\subfigure[Output ($r = 7$)]{
\includegraphics[width=0.22\textwidth]{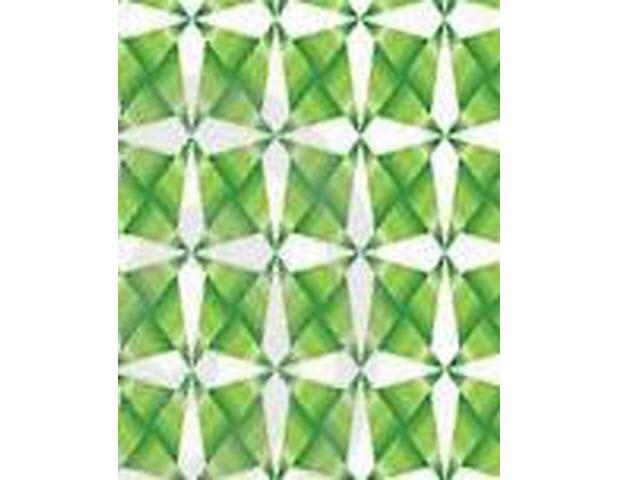}
}
\subfigure[Output ($r = 14$)]{
\includegraphics[width=0.22\textwidth]{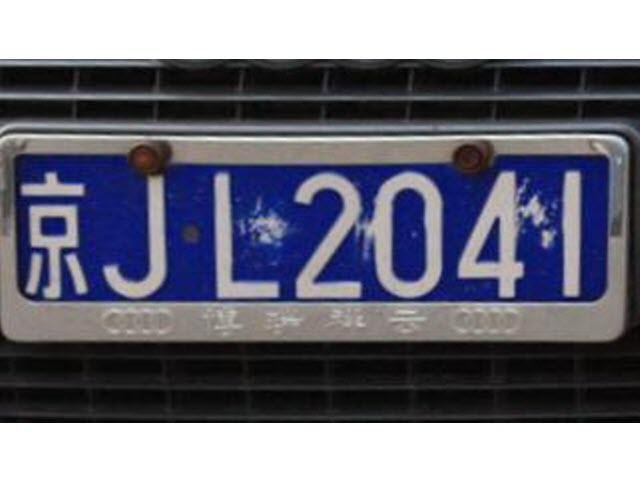}
}
}
\caption{\small \textbf{\small Representative Examples of Low-rank Textures and Our Results.} From left to right: an edge; a corner; a symmetric pattern, and a license plate. Top: deformed textures (high-rank as matrices); Bottom: the recovered low-rank representations.}
\label{fig:TILT-examples}
\end{figure*}

In this paper, we consider a 2D texture as a function $I^0(x,y)$, defined on $\mathbb{R}^2$. We say that $I^0$ is a {\em low-rank texture} if the family of one-dimensional functions $\{I^0(x, y_0)\mid y_0 \in \mathbb{R}\}$ span a finite low-dimensional linear subspace {\em i.e.},
\begin{equation}
r  \doteq \dim \big( \mbox{span}\{I^0(x, y_0) \mid y_0 \in \mathbb{R}\} \big) \le k
\label{eq:def}
\end{equation}
for some small positive integer $k$. If $r$ is finite, then we refer to $I^0$ as a rank-$r$ texture. Figure \ref{fig:TILT-examples} shows some ideal low-rank textures: a vertical or horizontal edge (or slope) can be considered as a rank-1 texture; and a corner can be considered as a rank-2 texture. To a large extent, the notion of low-rank texture unifies many of the conventional local features. By this definition, it is easy to see that {\em images of regular symmetric patterns always lead to low-rank textures}. Thus, the notion of low-rank texture encompasses a much broader range of ``features'' or regions than corners and edges.

Given a low-rank texture, obviously its rank is invariant under any scaling of the function, as well as scaling or translation in the $x$ and $y$ coordinates. That is, if $$g(x, y) \doteq c I^0(ax+ t_1, by + t_2)$$ for some constants $a, b, c \in \mathbb{R}_+,  t_1, t_2 \in \mathbb{R}$, then $g(x,y)$ and $I^0(x,y)$ have the same rank according to our definition  in \eqref{eq:def}. For most practical purposes, it suffices to recover any scaled or translated version of the low-rank texture $I^0(x,y)$, as the remaining ambiguity left in the scaling can often be easily resolved in practice by imposing additional constraints on the texture (see Section \ref{sec:implementation}). Hence, in this paper, unless otherwise stated, we view two low-rank textures {\em equivalent} if they are scaled and translated versions of each other: $$I^0(x,y) \sim c I^0 (ax + t_1, by + t_2),$$ for all $a, b, c \in \mathbb{R}_+, t_1, t_2 \in \mathbb{R}$. In homogeneous representation, this equivalence group consists of all elements of the form:
\begin{equation}
G \doteq \left\{
\left[ \begin{matrix} a & 0 & t_1 \\ 0 & b & t_2 \\ 0 & 0 & 1 \end{matrix}\right]  \in \Re^{3\times 3} \Big| \; a, b \in \Re_+, t_1, t_2 \in \Re
\right\}.
\end{equation}

In practice, we are never given the 2D texture as a continuous function in $\mathbb{R}^2$. Typically, we only have its values sampled on a finite discrete grid in $\mathbb{Z}^2$, of size $m\times n$ say. In this case, the 2D texture $I^0(x, y)$ is represented by an $m\times n$ real matrix. For a low-rank texture, we always assume that the size of the sampling grid is significantly larger than the intrinsic rank of the texture {\em i.e.,}\footnote{Notice that the scale of the window needs to be large enough to meet this assumption.}
$$ r \ll \min \{m, n\}. $$
It is easy to show that as long as the sampling rate is not one of the aliasing frequencies of the function $I^0$, the resulting matrix has the same rank as the continuous function.\footnote{In other words, the resolution of the image cannot be too low.} Thus, the 2D texture $I^0(x,y)$ when discretized as a matrix, also denoted by $I^0$ for convenience, has very low rank relative to its dimensions.

\begin{remark}[Low-rank Textures vs. Random Textures] \\ Conventionally, the word ``texture'' is used to describe image regions that exhibit certain spatially stationary stochastic properties ({\em e.g.,} grass, sand, fabrics). Such textures can be considered as random samples from a stationary stochastic process \citep{LevinaE2006-AS} and generally has full rank when viewed as a matrix. The low-rank ``textures'' defined here are complementary to such random textures. Here, low-rank textures correspond to regions in an image that have rather deterministic regular or periodic structures.
\end{remark}

\subsection{Deformed and Corrupted Low-rank Textures}
In practice, we typically never see a perfectly low-rank texture in a real image, largely due to two factors: 1. the change in viewpoint induces a transformation on the domain of the texture function; 2. the sampled values of the texture function are subject to many types of corruption such as quantization, noise, occlusions, etc. In order to correctly extract the intrinsic low-rank textures from such deformed and corrupted image measurements, we must first carefully model those factors and then seek ways to eliminate them.

\paragraph{Deformed Low-rank Textures.} Although many surfaces or structures in 3D exhibit low-rank textures, their images do not! Suppose that a low-rank texture $I^0(x,y)$ lies on a planar surface in the scene. The image $I(x, y)$ that we observe from a certain viewpoint is a transformed version of the original low-rank texture function $I^0(x,y)$:\footnote{By now, one should understand the reason of modeling low-rank texture as a function defined on a continuous domain $\Re^2$: we can talk about domain transformation freely. Any image or matrix representation of the texture is only a discrete sampling of this function. This allows us to generate transformed images of a low-rank texture by interpolating values of adjacent pixels.}
$$ I(x,y) = I^0 \circ \tau^{-1}(x,y) = I^0\left(\tau^{-1}(x,y)\right), $$
where $\tau : \mathbb{R}^2 \rightarrow \mathbb{R}^2$ belongs to a certain Lie group $\mathbb{G}$. In this paper, we assume $\mathbb{G}$ is either the rotation group $SO(2)$, or the 2D affine group $\mbox{Aff}(2)$, or the homography group $GL(3)$ acting linearly on the image domain.\footnote{Nevertheless, in principle, our method works for more general classes of domain deformations or camera projection models as long as they can be modeled well by a finite-dimensional parametric family.} In general, the transformed texture $I(x,y)$ as a matrix is no longer low-rank. For instance, a horizontal edge has rank one, but when rotated by $45^\circ$, it becomes a full-rank diagonal edge (see Figure \ref{fig:TILT-examples}(a)).

\paragraph{Corrupted Low-rank Textures.} In addition to domain transformations, the observed image of the texture might be corrupted by noise and occlusions or contain some pixels form the surrounding background. We can model such deviations as:
$$ I = I^0 + E$$
for some error matrix $E$. As a result, the image $I$ might no longer be a low-rank texture. In this paper, we assume that only a small fraction of the image pixels are corrupted by large errors, and hence, $E$ is a sparse matrix.

Our goal in this paper is to recover the exact low-rank texture $I^0$ from an image that contains a deformed and corrupted version of it. More precisely, we aim to solve the following problem:
\begin{problem}[Recovery of Low-rank Texture]
{\em Given a deformed and corrupted image of a low-rank texture: $I = (I^0 + E)\circ \tau^{-1}$, recover the low-rank texture $I^0$ and the domain transformation $\tau \in \mathbb{G}$. }
\end{problem}

The above formulation naturally leads to the following optimization problem:
\begin{equation}
\label{eqn:tilt} \min_{I^0,E,\tau} \;
\mathrm{rank}(I^0) + \gamma \|E\|_0 \quad \mathrm{s.t.} \quad I \circ \tau = I^0 + E,
\end{equation}
where $\|E\|_0$ denotes the number of non-zero entries in $E$.
That is, we aim to find the texture $I^0$ of the lowest possible rank and the error $E$ with the fewest possible nonzero entries that agrees with the observation $I$ up to a domain transformation $\tau$.  Here, $\gamma > 0$ is a weighting parameter that trades off the rank of the texture versus the sparsity of the error. For convenience, we refer to the solution $I^0$ found to this problem as a {\em Transform Invariant Low-rank Texture} (TILT).\footnote{By a slight abuse of terminology, we also refer to the procedure of solving the optimization problem as TILT.}


\begin{remark}[TILT vs. Affine-Invariant Features.] TILT is fundamentally different from the affine-invariant features or regions proposed in the literature (\cite{Affine-Scale-SIFT,Affine-SIFT}). Essentially, those features are extensions to SIFT features in the sense that their locations are very much detected in the same way as SIFT. The difference is that around each feature, an optimal affine transform is found that in some way ``normalizes'' the local statistics, say by maximizing the isotropy of the brightness pattern (\cite{Garding-1996}). Here TILT finds the best local deformation by minimizing the rank of the brightness pattern in a robust way. It works the same way for any image region of any size and for both affine and projective transforms (or even more general transformation groups that have smooth parameterization). More importantly, as we will see in Section \ref{sec:experiment}, our method is able to rectify all kinds of regions that are approximately low-rank ({\it e.g.} human faces, printed text) and the results match very well with human perception. Unlike SIFT features whose locations are difficult to predict or interpret by human vision, TILT has a nice WYSIWYG property:
\begin{quote}
``{\em What You See Is What You Get}.''
\end{quote}
\end{remark}

\begin{remark}[TILT vs. RASL.] We note that the optimization problem \eqref{eqn:tilt} is very similar to the robust image alignment problem studied in \cite{RASL}, known as RASL. This is because both RASL and TILT use the same mathematical framework (sparse and low-rank matrix decomposition with domain transformation) in their problem formulation. Although the formulation is similar, there are some important conceptual differences between the two problems. For instance, RASL treats each image as a vector and does not make use of any spatial structure within each image, whereas in this paper, TILT uses matrix rank and sparsity to study spatial structures within a 2D image. In this sense, RASL and TILT are highly complementary to each other: they try to capture temporal and spatial linear correlation in images, respectively. From an algorithmic point of view, TILT is simpler than RASL since it deals with only one image and one domain transformation whereas RASL deals with multiple images and multiple transformations, one for each image. We will propose many extensions to TILT to handle a wider range of textures and symmetries, most of which are not applicable to the image alignment problem that RASL strives to solve. Although beyond the scope of this paper, it remains to be seen in the future if one can combine TILT and RASL together to develop a richer class of tools for extracting more information from images.
\end{remark}

\begin{remark}[TILT vs. Transformed PCA.] One might argue that the low-rank objective can be directly enforced, as in {\em Transformed Component Analysis} (TCA) proposed by \cite{Frey1999-iccv}, which uses an EM algorithm to compute principal components, subject to domain transformations drawn from a known group. The TCA deals with Gaussian noise and essentially minimizes the 2-norm of the error term $E$. So the reader might wonder if such a ``transformed principal component analysis'' approach could apply to our image rectification problem here. Let us ignore gross corruption or occlusion for the time being. We could attempt to recover a rank-$r$ texture by solving the following optimization problem:
\begin{equation}
\label{eqn:tilt_l2}
\min_{I^0, \tau} \; \|I\circ\tau-I^0\|_F^2\quad \mathrm{s.t.}\quad \mathrm{rank}(I^0)\le r.
\end{equation}
One can solve \eqref{eqn:tilt_l2} by minimizing against the low-rank component $I^0$ and the deformation $\tau$ iteratively: with $\hat \tau$ fixed, estimate the rank-$r$ component $\hat I^0$ via PCA, and with $\hat I^0$ fixed, solve the deformation $\hat \tau$ in a greedy fashion to minimize the least-squares objective.\footnote{In fact, this simple iteration closely emulates the expectation-maximization (EM) procedure for solving the TCA problem proposed by \cite{Frey1999-iccv}.}

Figure \ref{fig:tilt_l2} shows some representative results of using such a ``Transformed PCA'' approach. However, even for simple patterns like the checker-board, it works only with a correct initial guess of the rank $r = 2$ beforehand. If we assume a wrong rank, say $r = 1$ or $3$, solving \eqref{eqn:tilt_l2} would not converge to a correct solution, even with a small initial deformation. For complex textures like a building facade shown in Figure \ref{fig:tilt_l2}, whose rank is impossible to guess in advance, we have to try all possibilities.  Moreover, \eqref{eqn:tilt_l2} can only handle small Gaussian noise. For images taken in real world, partial occlusion and other types of corruption are often present. The naive transformed PCA does not work robustly for such images. As we will see in the rest of this paper, the TILT algorithm that we propose next can automatically find the minimal matrix rank in an efficient manner and handle very large deformations and non-Gaussian errors of large magnitude.
\begin{figure}[t]
\centerline{
    \subfigure[$r=1$, fail]{
        \includegraphics[width=0.3\columnwidth]{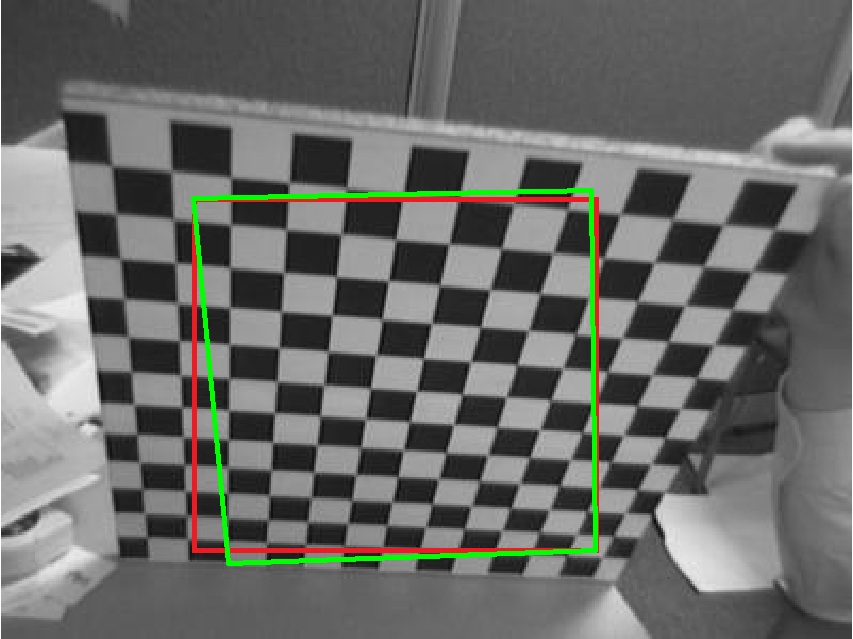}
    }
    \subfigure[$r=2$, succeed]{
        \includegraphics[width=0.3\columnwidth]{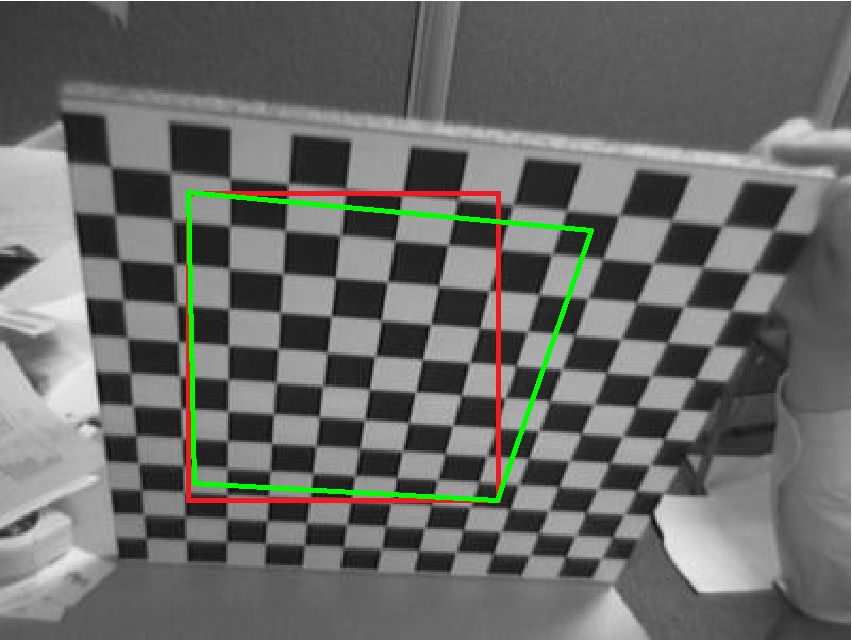}
    }
    \subfigure[$r=3$, fail]{
        \includegraphics[width=0.3\columnwidth]{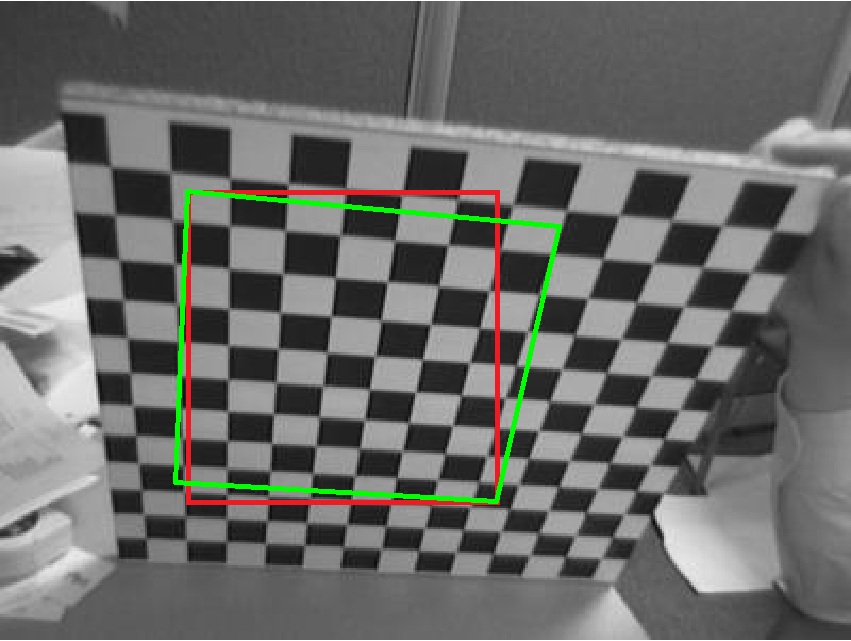}
    }
}

\centerline{
    \subfigure{
        \includegraphics[width=0.3\columnwidth]{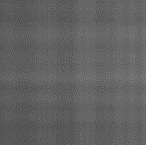}
    }
    \subfigure{
        \includegraphics[width=0.3\columnwidth]{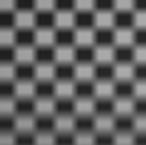}
    }
    \subfigure{
        \includegraphics[width=0.3\columnwidth]{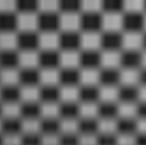}
    }
}

\centerline{
    \subfigure[$r=4$, fail]{
        \includegraphics[width=0.3\columnwidth]{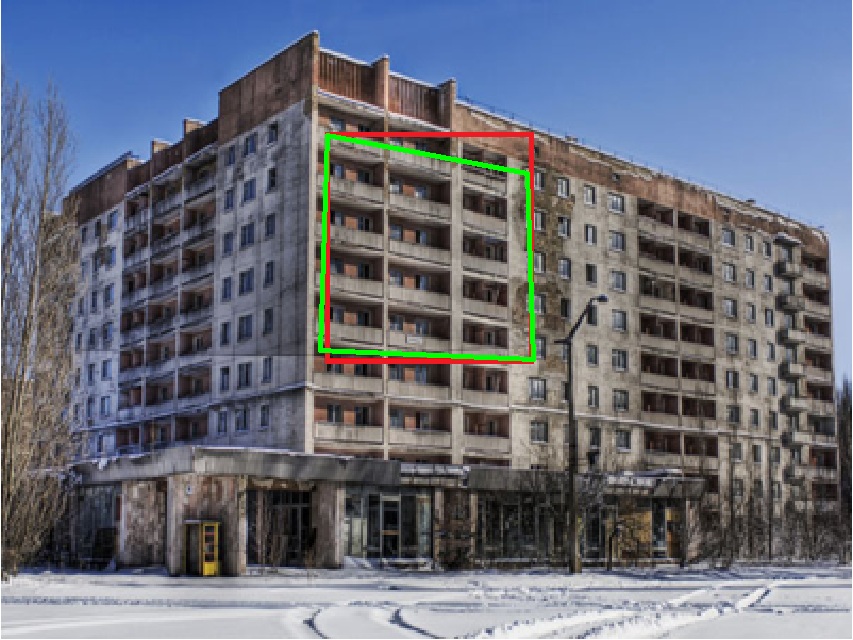}
    }
    \subfigure[$r=5$, succeed]{
        \includegraphics[width=0.3\columnwidth]{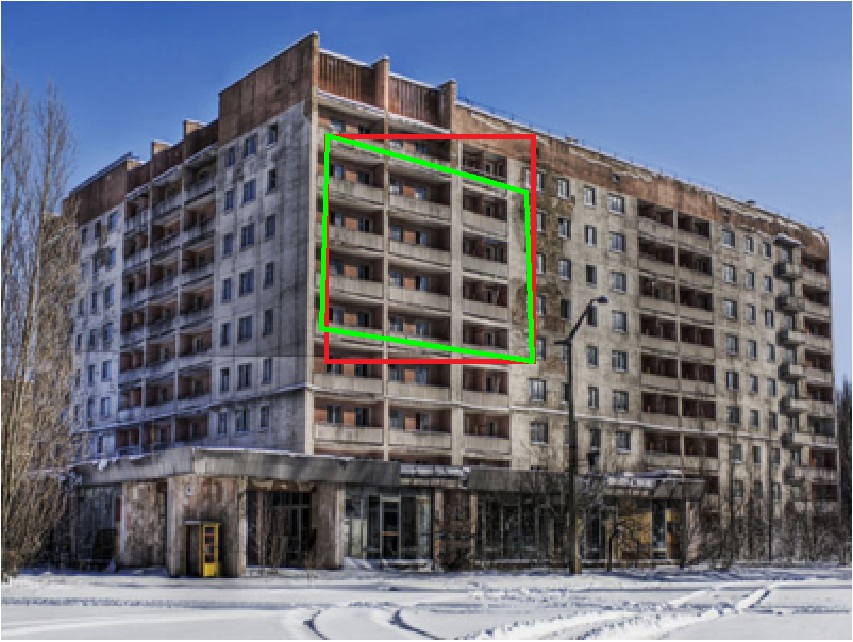}
    }
    \subfigure[$r=6$, fail]{
        \includegraphics[width=0.3\columnwidth]{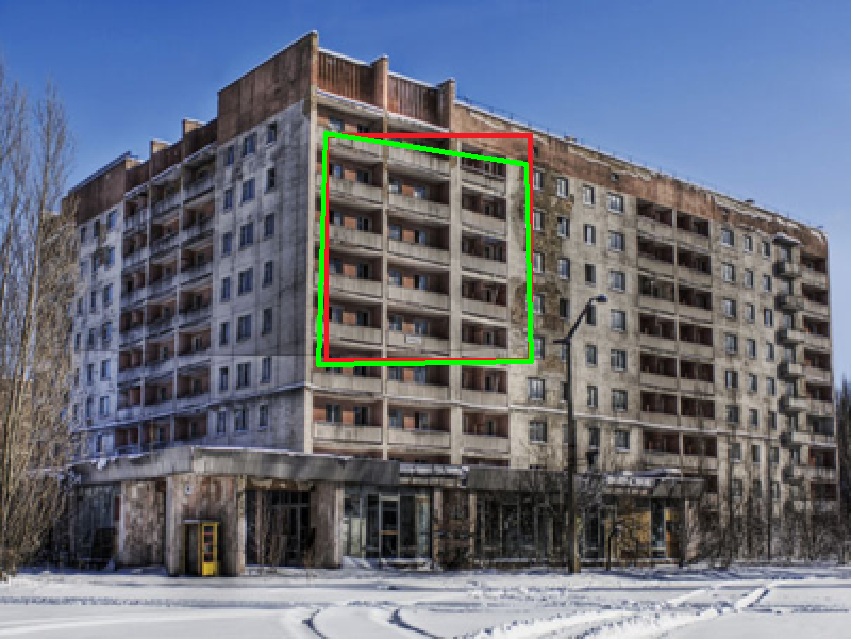}
    }
}

\centerline{
    \subfigure{
        \includegraphics[width=0.3\columnwidth]{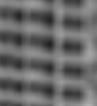}
    }
    \subfigure{
        \includegraphics[width=0.3\columnwidth]{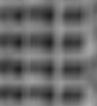}
    }
    \subfigure{
        \includegraphics[width=0.3\columnwidth]{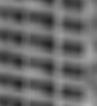}
    }
}

\caption{{\bf Transformed PCA:} Recovery of low-rank textures via solving \eqref{eqn:tilt_l2}. For a checker-board pattern if and only if we give the correct rank, $r = 2$, can we get correctly rectified textures. On a building facade, we try 6 different initial guesses of the rank from 1 to 6 and only rank $r = 5$ works \textit{approximately} well. }
\label{fig:tilt_l2}
\end{figure}

\end{remark}

\section{Solution by Iterative Convex Optimization}
\label{sec:solution}
As proposed in \citep{RASL}, although the rank function and the $\ell^0$-norm in the original problem \eqref{eqn:tilt} are extremely difficult to optimize (in general NP-hard), recent breakthroughs in sparse representation and low-rank matrix recovery have shown that under fairly broad conditions, they can be replaced by their convex surrogates \citep{Candes2009-pp,Chandrasekaran2009-pp}: the matrix {nuclear norm}\footnote{The nuclear norm of a matrix is the sum of all its singular values.} $\|I^0\|_*$ for $\mathrm{rank}(I^0)$ and the $\ell^1$-norm\footnote{The $\ell^1$-norm of a matrix is the sum of the absolute values of its entries.} $\|E\|_1$ for $\|E \|_0$, respectively. Thus, we end up with the following optimization problem:
\begin{equation}
\label{eqn:convex_tilt} \min_{I^0,E,\tau} \; \|I^0\|_* + \lambda
\|E\|_1 \quad \mathrm{s.t.} \quad I \circ \tau = I^0 + E.
\end{equation}

We note that although the objective function in the above problem is convex, the constraint $I \circ \tau = I^0 + E$ is nonlinear in $\tau \in \G$, and hence the problem is not convex. A common technique to overcome this difficulty is to linearize the constraint \citep{Baker2004-IJCV,RASL} around the current estimate and iterate. Thus, the constraint for the linearized version of our problem becomes
\begin{equation}
I \circ \tau +\nabla I  \Delta \tau  = I^0 + E,
\label{eqn:lin_cons}
\end{equation}
where  $\nabla I$ is the Jacobian: derivatives of the image w.r.t the transformation parameters.\footnote{Strictly speaking, $\nabla I$ is a 3D tensor: it gives a vector of derivatives at each pixel whose length is the number of parameters in the transformation $\tau$. When we ``multiply'' $\nabla I$ with another matrix or vector, it contracts in the obvious way which should be clear from the context.} The optimization problem in \eqref{eqn:convex_tilt} reduces to
\begin{equation}
\label{eqn:convex_linear_tilt} \min_{I^0,E,\Delta\tau} \; \|I^0\|_* + \lambda
\|E\|_1 \quad \mathrm{s.t.} \quad I \circ \tau +\nabla I  \Delta \tau  = I^0 + E.
\end{equation}
The linearized problem above is a convex program and is amenable to efficient solution. Since the linearization is only a local approximation to the original nonlinear problem, we solve it iteratively in order to converge to a (local) minimum of the original non-convex problem \eqref{eqn:convex_tilt}. The algorithm has been summarized as Algorithm \ref{alg:outer_loop}.

\begin{algorithm}[t]
\caption{{\bf (The TILT Algorithm)}}%
\begin{algorithmic}
\STATE \textbf{INPUT:} Input image $I \in
\mathbb{R}^{w\times h}$, initial transformation $\tau \in \mathbb{G}$ (affine or projective), and a weight $\lambda > 0$.
\STATE \textbf{WHILE} not converged \textbf{DO} \\
\STATE \qquad \textbf{Step 1:} Normalization and compute Jacobian:
$$
I \circ \tau \leftarrow \frac{I \circ \tau}{\|I\circ \tau\|_F}; \quad \nabla I \leftarrow
\frac{\partial}{\partial \zeta}\left(\frac{\vec(I \circ \zeta) }{\|\vec(I \circ \zeta)\|_F}\right) \Bigr|_{\zeta = \tau};
$$
\qquad \textbf{Step 2 (inner loop):} Solve the linearized problem:
$$
\begin{array}{ccl}
(I^{0*},E^*, \dt^*) & \leftarrow & \arg\min_{I^0,E,\dt} \;\; \|I^0\|_* + \lambda \|E\|_1 \\
& &\quad \mathrm{s.t.} \;\; I \circ \tau + \nabla I \Delta \tau = I^0 + E;
\end{array}
$$
\qquad \textbf{Step 3:} Update the transformation: $\tau \leftarrow \tau + \dt^*$;
\STATE \textbf{END WHILE}
\STATE \textbf{OUTPUT:} Optimal solution $I^{0*}$, $E^*$, $\tau^*$ to problem \eqref{eqn:convex_tilt}.
\end{algorithmic}
\label{alg:outer_loop}
\end{algorithm}

The iterative linearization scheme outlined above is a common technique in optimization to solve nonlinear problems. It can be shown that this kind of iterative linearization converges quadratically to a local minimum of the original non-linear problem. A complete proof is out of the scope of this paper. We refer the interested reader to \cite{Peng2010-pp,Cromme1978-NM,Jittorntrum1980-NM} and the references therein.


\subsection{Fast Algorithm Based on Augmented Lagrange Multiplier Methods}
\label{sec:alm}
The most computationally expensive part of Algorithm \ref{alg:outer_loop} is solving the convex program in the inner loop (Step 2) of Algorithm \ref{alg:outer_loop}. This can be cast as a semidefinite program and can be solved using conventional algorithms such as interior-point methods. While interior-point methods have excellent convergence properties, they do not scale very well with problem size and hence, unsuitable for real applications involving large images. Fortunately, there has been a recent flurry of work in developing fast, scalable algorithms for nuclear norm minimization \citep{Cai2010-SIAM,Toh2009-PJO,Ganesh2009-CAMSAP,Lin2009-2-pp}. To solve the linearized problem in \eqref{eqn:convex_linear_tilt}, we use the Augmented Lagrange Multiplier (ALM) method \citep{Bertsekas2004,Lin2009-2-pp}. For the sake of completeness, in this section we explain how the ALM method can be adapted to solve our problem, and also comment on some implementation details for improving stability and range of convergence.

\subsubsection{General Formulation of ALM}

We first review the ALM algorithm in a more general setting, rather than for our specific problem. This will be useful later when we deal with  different variations of the TILT algorithm that can all be solved under the same algorithmic framework described here.

Let us consider convex optimization problems of the form:
\begin{equation}
\min_X \, f(X) \quad \mathrm{s.t.}\quad  \mathcal{A}(X) = b,
\label{eqn:gen_prob}
\end{equation}
where $f$ is a convex (not necessarily smooth) function, $\mathcal{A}$ is a linear function, and $b$ is a vector of appropriate dimension. The basic idea of Lagrangian methods is to convert the above constrained optimization problem into an unconstrained problem that has the same optimal solution.

For the above problem \eqref{eqn:gen_prob}, we define the augmented Lagrangian function as follows:
\begin{equation}
\mathcal{L}_\mu(X,Y) = f(X) + \langle Y, b-\mathcal{A}(X)\rangle + \frac{\mu}{2}\|b-\mathcal{A}(X)\|_2^2,
\label{eqn:gen_lag}
\end{equation}
where $Y$ is a Lagrange multiplier vector of appropriate dimension, $\|\cdot\|_2$ denotes the Euclidean norm, and $\mu > 0$ denotes the penalty imposed upon infeasible points. The following result from \cite{Bertsekas2004} establishes an important relation between the original problem \eqref{eqn:gen_prob} and its augmented Lagrangian function \eqref{eqn:gen_lag}.

\begin{theorem}[Optimality of ALM]
Suppose that $\hat{X}$ is the optimal solution to \eqref{eqn:gen_prob}. Then, for appropriate choice of $Y$ and sufficiently large $\mu$, we have
$$
\hat{X} = \arg\min_X \, \mathcal{L}_\mu(X,Y).
$$
\end{theorem}
Thus, we could solve an unconstrained convex minimization problem in order to obtain the solution to the convex program \eqref{eqn:gen_prob}. This result, while of theoretical interest, is not directly useful in practice since the choice of $Y$ and $\mu$ is not known a priori.

ALM methods are a class of algorithms that simultaneously minimize the augmented Lagrangian function and compute an appropriate Lagrange multiplier. The basic ALM iteration proposed in \cite{Bertsekas2004} is given by
\begin{equation}
\begin{array}{cll}
X_{k+1} & = & \arg\min_X \, \mathcal{L}_{\mu_k}(X,Y_k), \\
Y_{k+1} & = & Y_k + \mu_k \,\left(b-\mathcal{A}(X_k)\right), \\
\mu_{k+1} & = & \rho\cdot\mu_k,
\end{array}
\label{eqn:gen_alm}
\end{equation}
where $\{\mu_k\}$ is a monotonically increasing positive sequence ($\rho > 1$). Thus, we have reduced the original optimization problem \eqref{eqn:gen_prob} to a sequence of unconstrained convex programs.

The above iteration is computationally useful only if $\mathcal{L}_\mu(X,Y)$ is easy to minimize with respect to $X$. For the problems encountered in this paper, this turns out to be the case indeed. This can be attributed to the following key property of the matrix nuclear norm and 1-norm:
\begin{equation}
\begin{array}{ccl}
\mathcal{S}_\mu(Y_1+Y_2)\! &\! = & \! \arg\min_X \mu \|X\|_1 \!-\! \langle X,Y_1\rangle  \!+  \frac{1}{2}\|X\!-\!Y_2\|_F^2, \\
U\mathcal{S}_\mu[\Sigma]V^*\! &\! = &\! \arg\min_X  \mu \|X\|_* \!-\! \langle X,\!W_1\rangle \!+\! \frac{1}{2}\|X\!-\!W_2\|_F^2,
\end{array}
\end{equation}
where $U\Sigma V^*$ is the Singular Value Decomposition (SVD) of $(W_1+W_2)$, and $\mu$ is any non-negative real constant. Here, $\mathcal{S}[\cdot]$ represents the soft-thresholding or \emph{shrinkage} operator which is defined on scalars as follows:
\begin{equation}
S_\mu[x] = \sign(x)\cdot\left(|x|-\mu\right),
\end{equation}
where $\mu \geq 0$. The shrinkage operator is extended to vectors and matrices by applying it elementwise. We now discuss how this iterative scheme can be applied to our linearized convex program \eqref{eqn:convex_linear_tilt}.

\subsubsection{Solving TILT by Alternating Direction Method}

For the problem given in \eqref{eqn:convex_linear_tilt}, the augmented Lagrangian is defined as:
\begin{eqnarray}
L_\mu(I^0, E, \Delta \tau, Y) & \doteq & f(I^0,E) +  \langle Y,  R(I^0, E, \Delta \tau) \rangle \nonumber \\ &&+ \frac{\mu}{2} \left\| R(I^0, E, \Delta \tau) \right\|_F^2,
\label{eqn:lagrangian}
\end{eqnarray}
where $\mu > 0$, $Y$ is a Lagrange multiplier matrix, $\langle \cdot,\cdot\rangle$ denotes the matrix inner product, and
\begin{equation*}
\begin{array}{ccl}
f(I^0,E) & = & \|I^0\|_* + \lambda \|E\|_1, \\
R(I^0, E, \Delta \tau) & = & I \circ \tau +\nabla I \Delta \tau  - I^0 - E.
\end{array}
\end{equation*}

From the above discussion, the basic ALM iteration scheme for our problem is given by
$$
\begin{array}{rcl}
 (I^0_{k}, E_{k}, \Delta \tau_{k}) & = & \arg\min_{I^0, E, \Delta \tau} L_{\mu_k}(I^0, E, \Delta \tau, Y_{k-1}),  \\
 Y_{k} & = & Y_{k-1} + \mu_{k-1} R(I^0_k, E_k, \Delta \tau_k).
\end{array}
$$
Throughout the rest of the paper, we will always assume that $\mu_k = \rho^k\,\mu_0$ for some $\mu_0 > 0$ and $\rho > 1$, unless otherwise specified.

We now focus on efficiently solving the first step of the above iterative scheme. In general, it is computationally expensive to minimize over all the variables $I^0$, $E$ and $\Delta\tau$ simultaneously. So, we adopt a common strategy to solve it {\em approximately} by adopting an {\em alternating minimizing} strategy {\it i.e.,} minimizing with respect to $I^0$, $E$ and $\Delta\tau$ one at a time:
\begin{equation}
\begin{array}{rcl}
I^0_{k+1} &= &  \arg\min_{I^0} L_{\mu_k}(I^0, E_k, \Delta \tau_k, Y_{k}), \\
E_{k+1} &= & \arg\min_{E} L_{\mu_k}(I^0_{k+1}, E, \Delta \tau_k, Y_{k}), \\
\Delta\tau_{k+1} &= & \arg\min_{\Delta \tau} L_{\mu_k}(I^0_{k+1}, E_{k+1}, \Delta \tau, Y_{k}).
\end{array}
\label{eqn:optimizations}
\end{equation}
Due to the special structure of our problem, each of the above optimization problems has a simple closed-form solution, and hence, can be solved in a single step. More precisely, the solutions to \eqref{eqn:optimizations} can be expressed explicitly using the shrinkage operator as follows:
\begin{equation}
\begin{array}{rcl}
I^0_{k+1} &\; \leftarrow\; &  U_k\mathcal{S}_{\mu_k^{-1}} [\Sigma_k]V^T_k,   \\
E_{k+1} &\; \leftarrow\; & \mathcal{S}_{\lambda \mu_k^{-1}} [I \circ \tau +\nabla I \Delta \tau_{k}  - I^0_{k+1} + \mu_k^{-1}Y_k ],\\
\Delta\tau_{k+1} &\; \leftarrow\;  & (\nabla I)^\dag (- I \circ \tau  + I^0_{k+1} + E_{k+1} - \mu_k^{-1}Y_k),
\end{array}
\end{equation}
where $U_k\Sigma_k V^T_k$ is the SVD of $\big( I \circ \tau  +\nabla I \Delta \tau_{k} - E_{k} + \mu_k^{-1}Y_k\big)$, and $(\nabla I)^\dag$ denotes the Moore-Penrose pseudo-inverse of $\nabla I$.

From experiments, we observe that the above algorithm is much faster than all other alternative convex optimization schemes (such as the interior point method, accelerated proximal gradient, etc.). Although the convergence  of the ALM method \eqref{eqn:gen_alm} has been well established in the optimization literature,  its approximation by the above alternating minimization, known as {\em alternating direction method (ADM) of multipliers}, is not always guaranteed to converge to the optimal solution. If there are only two alternating terms, its convergence has been well-studied and established \citep{Glowinski1975-TR,Gabay1976-CMA,Eckstein1992-MP}. Somewhat surprisingly, however, very little is proven for the convergence of cases where there are more than three alternating terms, despite overwhelming empirical success with such schemes. Recently, \cite{Yuan2010-pp} obtained a convergence result for a certain three-term alternation applied to the noisy principal component pursuit problem (see also \citep{He2009-COA}). However, the scheme proposed and proved in \citep{Yuan2010-pp} is slightly different from the direct ADM scheme \eqref{eqn:optimizations} and is much slower in practice. The convergence of the ADM scheme \eqref{eqn:optimizations} remains an open problem although in practice it gives the simplest and fastest algorithm.

We summarize the ADM scheme for solving \eqref{eqn:convex_linear_tilt} as Algorithm \ref{alg:inner_loop}. We choose the sequence $\{\mu_k\}$ to satisfy $\mu_{k+1} = \rho\,\mu_k$ for some $\rho > 1$. We note that the operations in each step of the algorithm are very simple with the SVD computation being the most computationally expensive step.\footnote{Empirically, we notice that for larger window sizes (over $100\times 100$ pixels), it is much faster to run the partial SVD instead of the full SVD, if the rank is known to be very low.}

\begin{algorithm}[t]
\caption{{\bf (Solving Inner Loop of TILT)}}%
\begin{algorithmic}
\STATE \textbf{INPUT:} The current (deformed and normalized) image $I \circ \tau \in \mathbb{R}^{m\times n}$ and its Jacobian $\nabla I$ against deformation $\tau$, and $\lambda > 0$.
\STATE {\bf Initialization:} $k = 0, Y_0 = 0, E_0 = 0, \Delta \tau_0 = 0, \mu_0 > 0, \rho > 1$;
\STATE {\bf WHILE} not converged {\bf DO}
\STATE $\qquad (U_k,\Sigma_k,V_k) = \mathrm{svd}\big( I \circ \tau  +\nabla I \Delta \tau_{k} - E_{k} + \mu_k^{-1}Y_k\big)$;
\STATE $\qquad I^0_{k+1} =  U_k\mathcal{S}_{\mu_k^{-1}} [\Sigma_k]V^T_k$;
\STATE $\qquad E_{k+1}  =   \mathcal{S}_{\lambda \mu_k^{-1}} [I \circ \tau +\nabla I \Delta \tau_{k}  - I^0_{k+1} + \mu_k^{-1}Y_k ]$;
\STATE $\qquad \Delta\tau_{k+1}  =  (\nabla I)^\dag (- I \circ \tau  + I^0_{k+1} + E_{k+1} - \mu_k^{-1}Y_k)$;
\STATE $\qquad Y_{k+1} = Y_k + \mu_k (I \circ \tau + \nabla I \Delta\tau_{k+1} - I^0_{k+1} - E_{k+1})$;
\STATE $\qquad \mu_{k+1} = \rho \mu_k$;
\STATE {\bf END WHILE}
\STATE \textbf{OUTPUT:} solution $(I^0$, $E$, $\dt)$ to problem \eqref{eqn:convex_linear_tilt}.
\end{algorithmic}
\label{alg:inner_loop}
\end{algorithm}

\subsection{Implementation Details}\label{sec:implementation}

In the previous section, we described how the {\em linearized} and {\em convexified} TILT problem \eqref{eqn:convex_tilt} can be solved efficiently using the ALM algorithm. However, there are a few caveats in applying it to real images. In this section, we discuss some possible ways to deal with these issues and make the problem well-defined. We also discuss some specific implementation details that could potentially improve the range of convergence of our algorithm.

\paragraph{Constraints on the Transformations.}
As discussed in Section \ref{sec:formulation}, there are certain ambiguities in the definition of low-rank texture. The rank of a low-rank texture function is invariant with respect to scaling in the pixel values, scaling in each of the coordinate axes, and translation along any direction. Thus, in order for the problem to have a unique, well-defined optimal solution, we need to eliminate these ambiguities. In Step 1 of Algorithm \ref{alg:outer_loop}, the intensity of the image is renormalized in each iteration in order to eliminate the ambiguity of scale in the pixel values. Otherwise, the algorithm may tend to converge to a ``globally optimal'' solution by zooming into a black pixel or dark region of the image.

To deal with the ambiguities in the domain transformation, we could add some additional constraints to the problem. Let $\tau(\cdot)$ represent the transformation. Suppose that the support of the initial image window $\Omega$ is a rectangle (call the edges $e_1$ and $e_2$) with the length of the two edges being $L(e_1) = a$ and $L(e_2) = b$, so that the total area $S(\Omega) = ab$.

For affine transformations, to eliminate the ambiguity in translation, we typically enforce that the center $x_0$ of the initial rectangular region $\Omega$ remain fixed before and after the transformation {\it i.e.,} $\tau (x_0) = x_0$. This imposes a set of linear constraints on $\Delta \tau$ of the form:
\begin{equation}
A_t \Delta \tau = 0.
\label{eqn:cons1}
\end{equation}
To eliminate the ambiguities in scaling the coordinates, we enforce (only for affine transformations) that the area and the ratio of edge length remain constant before and after the transformation, i.e. $S(\tau(\Omega)) = S(\Omega)$ and $L(\tau(e_1))/L(\tau(e_2)) = L(e_1)/L(e_2)$.
In general, these equalities impose additional nonlinear constraints on the desired transformation $\tau$ in problem \eqref{eqn:convex_tilt}. Similar to the way we dealt with the non-linearity in the constraint in \eqref{eqn:convex_tilt}, we can linearize these additional constraints w.r.t. the transformation parameters $\tau$ and obtain another set of linear constraints on $\Delta \tau$ denoted by:
\begin{equation}
A_s \Delta \tau = 0.
\label{eqn:cons2}
\end{equation}
We have given a more detailed explanation and derivation of these two sets of linear constraints in the appendix.

For projective transformations, we typically fix two points,\footnote{In fact, one can use the same set of constraints as the affine case. But from our experience, the algorithm is more stable with the initialization of two points. In addition, as we will explain, the parameterization is more geometrically meaningful.} the two diagonal corners of the initial rectangular window or of the parallelogram if initialized with the result of the affine TILT.\footnote{In practice, we almost always initialize the projective case with the result from the affine case.} Notice that a homography matrix has a total of eight degrees of freedom. If the low-rank texture is associated with certain symmetric pattern that has two sets of parallel lines, the x and y-axes of the rectified low-rank texture then correspond to the two vanishing points. The two vanishing points and the two fixed points together impose exactly eight constraints and uniquely determine the homography. Hence, with this parameterization, there is no ambiguity in the optimal solution.

Thus, to eliminate the scaling and translation ambiguities in the solution, we simply add a set of linear constraints to the optimization problem \eqref{eqn:convex_linear_tilt}. The resulting convex program can be solved again using the ALM algorithm. This would involve making very small modifications to Algorithm \ref{alg:inner_loop} to incorporate the additional linear constraints.\footnote{We only have to introduce an additional set of Lagrangian multipliers and then revise accordingly the update equation associated with $\Delta \tau_{k+1}$.}

\paragraph{Multi-Resolution Approach.} While the above formulation works reasonably well in practice, the presence of arbitrarily shaped sharp features or contours on an otherwise smooth low-rank texture can cause the TILT algorithm to converge to a local minima that is not the desired solution. Hence, to cope with large deformations, we adopt a multi-resolution approach. This is a common technique in many computer vision algorithms wherein we construct a pyramid of images, starting from the input image, by subsequently blurring and downsampling it. The problem is then solved at the lowest resolution first. The solution thus obtained is used to initialize the algorithm at the adjacent level of higher resolution, and this procedure is repeated for all levels. In practice, the multi-resolution approach not only improves the range of transformations that our algorithm can handle, but it also improves the running time of the algorithm significantly. This is because, the convex programs can be solved much faster at the lower resolutions, and since the initialization at the higher resolution is better, the number of iterations to convergence is typically very small (less than 20).

An important consideration while incorporating the multi-resolution approach for the TILT algorithm is the fact that the convex relaxation discussed in Section \ref{sec:solution} is tight only at higher dimensions.\footnote{The convex relaxation has a failure probability associated with it which typically decays as $O(n^{-\alpha})$, for some $\alpha > 0$, assuming that the matrices involved have size $n \times n$.} Although it is very difficult to analytically estimate the minimum optimal size of the image, in practice, we find that our method works well for windows of size larger than $20 \times 20$ pixels. In our implementation, we use a Gaussian kernel to blur the image and consider up to two levels of downsampling, each by a factor of 2 w.r.t. its adjacent higher level of resolution. We also ensure that the size of the image at the lowest resolution is at least $20 \times 20$ pixels. We tested the speed of this scheme in Matlab on a 3Ghz PC. Fixing the initial window to have size $50\times 50$, the running time is less than 6 seconds, averaged over 100 trials.


\paragraph{Branch-and-Bound Scheme.} We can increase the range of deformation that our algorithm can handle significantly by employing a branch-and-bound scheme. For instance, in the affine case, we initialize Algorithm \ref{alg:outer_loop} with different deformations (e.g., a combination search for all 4 degrees of freedom for affine transforms with no translation). Any affine transformation can be parameterized by $[A \;\; b]\in \R^{2\times 2} \times \R^2$. Since we fix the center of the window, we effectively set $b = 0$. The remaining 4 parameters of the transformation denote the scaling along the $x$ and $y$-axes, rotation, and skew. As discussed in Section \ref{sec:formulation}, the scaling along the canonical axes does not change the rank of the texture, and hence, we ignore the ambiguity in it. Thus, we are left with two parameters - skew and rotation - that need to be determined. In other words, we can parameterized the affine matrix $A$ as: $$A(\theta, t)=\left[\begin{matrix} \cos{\theta} & -\sin{\theta} \\ \sin{\theta} & \cos{\theta} \end{matrix}\right]\times \left[\begin{matrix} 1 & t \\ 0 & 1 \end{matrix}\right].$$  We partition the parameter space (rotation and skew) into multiple regions and perform a greedy search on the regions one-by-one. We first run TILT for various initializations of the rotation angle. We choose the one that minimizes the cost function, and use this as an initialization to search for the skew parameters along the $x$-direction first, and subsequently along the $y$-direction. The parameters that minimize the cost function is the output of the branch-and-bound scheme.

A natural concern about such a branch-and-bound scheme is its effect on speed. Within the multi-resolution scheme, we only need to perform branch-and-bound at the level of lowest resolution, find the best solution, and use it to initialize the higher-resolution levels. Since Algorithm \ref{alg:outer_loop} is extremely fast for small matrices at the lowest-resolution level, running multiple instances with different initializations does not significantly affect the overall speed. In a similar spirit, to find the optimal projective transform (homography), we always find the optimal affine transform first and then use it to initialize the algorithm.\footnote{Notice that, for a perspective image of a plane, the affine model is approximately true if the size of the patch is small compared to the distance. The projective model however applies regardless of the size.} From our experience, we found that with such initialization, we normally do not have to use the branch-and-bound scheme for the projective transformation case.

\section{Experimental Results}
\label{sec:experiment}
In this section, we present the results of the proposed TILT algorithm on various natural and artificial low-rank textures. We first present some results quantifying the performance range of our algorithm. We then present examples from many different categories of natural images where TILT can recover the inherent symmetrical texture in the images. Finally, we present some examples where TILT does not recover the low-rank texture and examine the reasons for such failures.

\subsection{Range of Convergence of TILT}
For most low-rank textures, the proposed Algorithm \ref{alg:outer_loop} has a fairly large range of convergence, {\em without using any branch-and-bound}. In this section, we give a careful characterization of the range of convergence (ROC) of the proposed algorithm on a standard checker-board pattern.

\begin{figure*}[!ht]
\centering{
    \subfigure[Representative Input Images in Each Region]{
        \includegraphics[width=0.82\columnwidth]{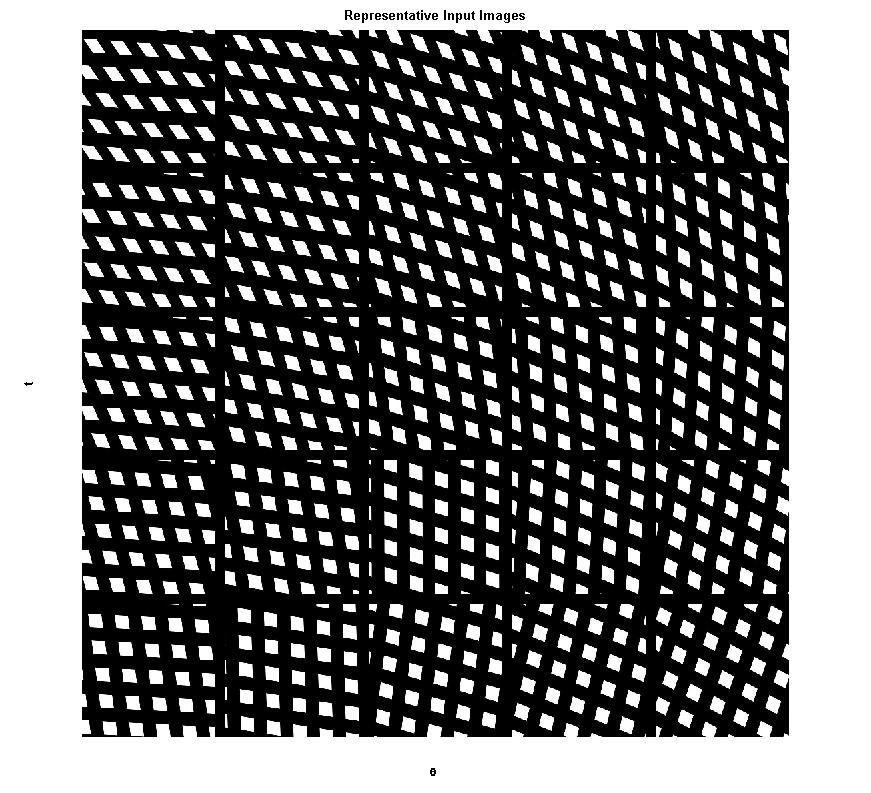}
    }
    \subfigure[Convergence Probability Map]{
        \includegraphics[width=0.97\columnwidth]{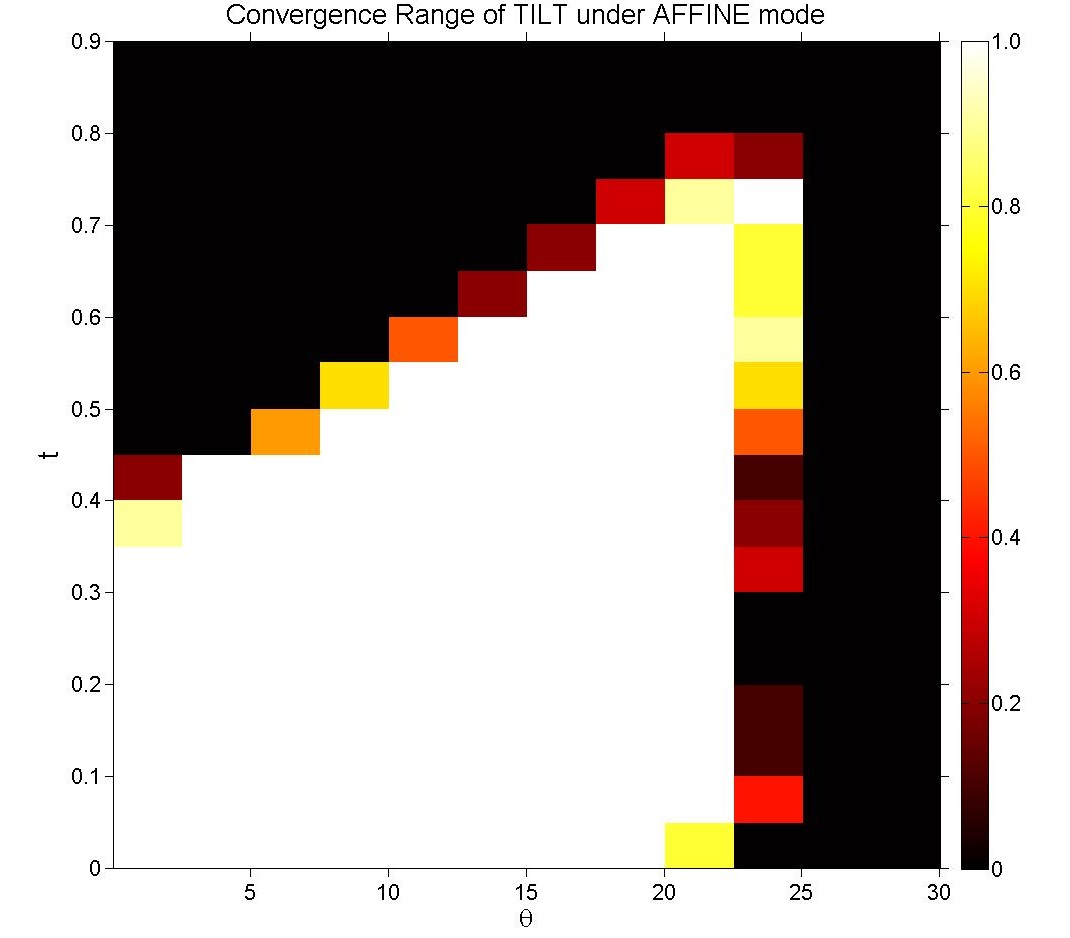}
    }
}
\caption{{\bf Range of Convergence for Affine Transform} without branch-and-bound. Image on the left: initial input images that correspond to different regions of the range of parameter space in the plot on the right. Plot on the right: $x$-axis: rotation angle $\theta$; $y$-axis: skew parameter $t$. White region indicates success in all trials while the black region indicates failure in all trials.}
\label{fig:convergence}
\end{figure*}

\paragraph{Affine Case.} We deform a checker-board like pattern by a wide range of affine transforms of the form: $y=Ax+b, x, y\in \R^2$, and test if the algorithm converges back to the correct solution. We parameterize the affine matrix $A$ as $A(\theta, t)=\left[\begin{smallmatrix} \cos{\theta} & -\sin{\theta} \\ \sin{\theta} & \cos{\theta} \end{smallmatrix}\right]\times \left[\begin{smallmatrix} 1 & t \\ 0 & 1 \end{smallmatrix}\right].$
We change $(\theta, t)$ within the range $\theta\in[0, \pi/6]$ with step size $\pi/60$, and $t\in[0, 1]$ with step size $0.05$. We repeat the simulations 10 times in each region and compute the success rate. Figure \ref{fig:convergence}(b) shows the rate of success for all regions. Notice that the algorithm always finds the correct solution for up to $\theta = 20^\circ$ of rotation and skew (or warp) of up to $t = 0.4$. We note that, due to its rich symmetries and sharp edges, the checker-board like pattern is a challenging case for ``global'' convergence as at many angles, its image corresponds to a local minimum that has relatively low rank. In practice, we find that for most symmetric patterns in urban scenes (as shown in Figure \ref{fig:geometry}), our algorithm converges for a much larger range without any branch-and-bound.
So the large ROC ensures that a simple partition of the parameter space with a branch-and-bound scheme can make the TILT algorithm work for the entire range of affine transformations.

\paragraph{Projective Case.} For the case of projective transformations (homographies), even if we fix two points, there are still four remaining degrees of freedom. It is difficult to illustrate the range of convergence for all four dimensions together. So here we test the range of convergence for some of the most representative projective transformations that we normally encounter in real-world images: a planar low-rank pattern rotating in front of a perspective camera.

\begin{figure*}[th]
\centering{
\subfigure{\includegraphics[width=0.77\columnwidth]{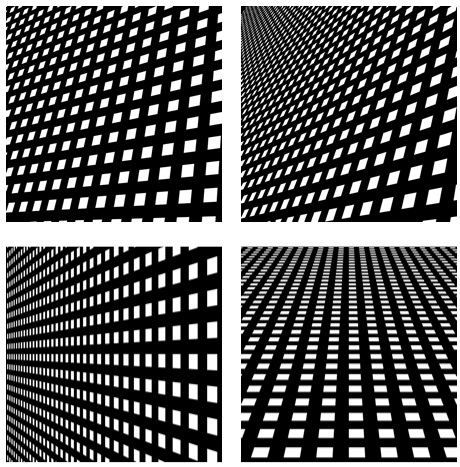}}
\subfigure{\includegraphics[width=1.07\columnwidth]{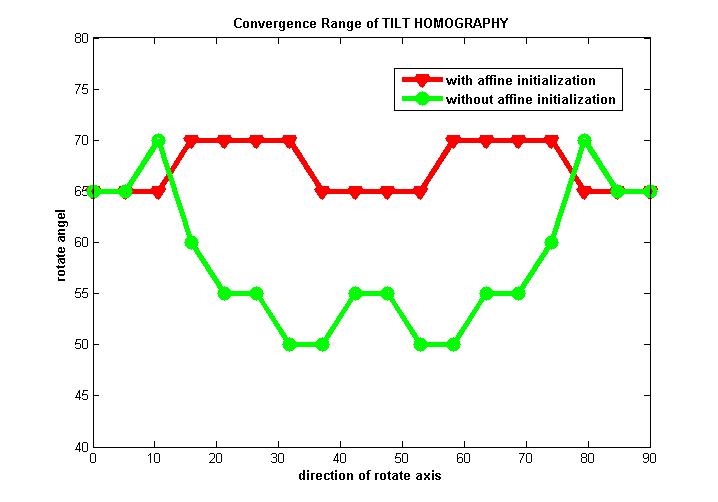}}}
\caption{{\bf Range of Convergence for Projective Transform}. Image on the left: Representative initial input images for which the TILT algorithm succeeds without any special initialization or branch and bound.  Plot on the right: $x$-axis: the position of the rotation axis; $y$-axis: the amount of rotation. Green curve: without initialization with affine TILT. Red curve: initialized with affine TILT.}
\label{fig:convergence2}
\end{figure*}

More specifically, we put a standard checker-board pattern in front of a standard perspective camera -- the image plane is the $xy$-plane and the optical axis is the $z$-axis. We rotate the pattern along a line through the origin within the $xy$-plane. We indicate the location of the axis of rotation by the angle it makes with the $x$-axis. Experimentally, we find the limits of the TILT algorithm by gradually increasing the amount of rotation along each axis (from $0^\circ$ to $90^\circ$ at a step of $5^\circ$). We also change the rotation axis from the $x$-direction ($0^\circ$) to the $y$-direction ($90^\circ$).\footnote{The setting is symmetric and the pattern is symmetric so we only have to verify the range of convergence for the first quadrant.} Figure \ref{fig:convergence2} shows the range of convergence of TILT under this setting. The curves indicate when TILT fails for the first time, or in other words, TILT succeeds for all cases below the curves.

The two curves in the plot compare two cases: the first case (green curve) is just the basic projective TILT without any special initialization nor any branch and bound; the second case (red curve) is the projective TILT initialized with the results from the affine TILT. From these results, we may conclude:
\begin{itemize}
\item The basic projective TILT works extremely well for the slanted checker-board like pattern -- it converges up to $50^\circ$ of rotation in all directions.
\item Initialization with the affine TILT normally boosts the range of convergence for the projective TILT up to $65^\circ$ or rotation, in some cases increasing by as much as $20^\circ$.
\end{itemize}
There are many possible ways to further improve the range of convergence for the TILT algorithm. So far, we have always used a square window as the initial window. As we will see with experiments in later sections, TILT could work much better if the initial window is chosen in a way that is more adaptive to the orientation of the texture as well as the scale of the texture.

\subsection{Robustness of TILT}
\begin{figure*}[hbt]
\centering{
\includegraphics[width=0.75\columnwidth]{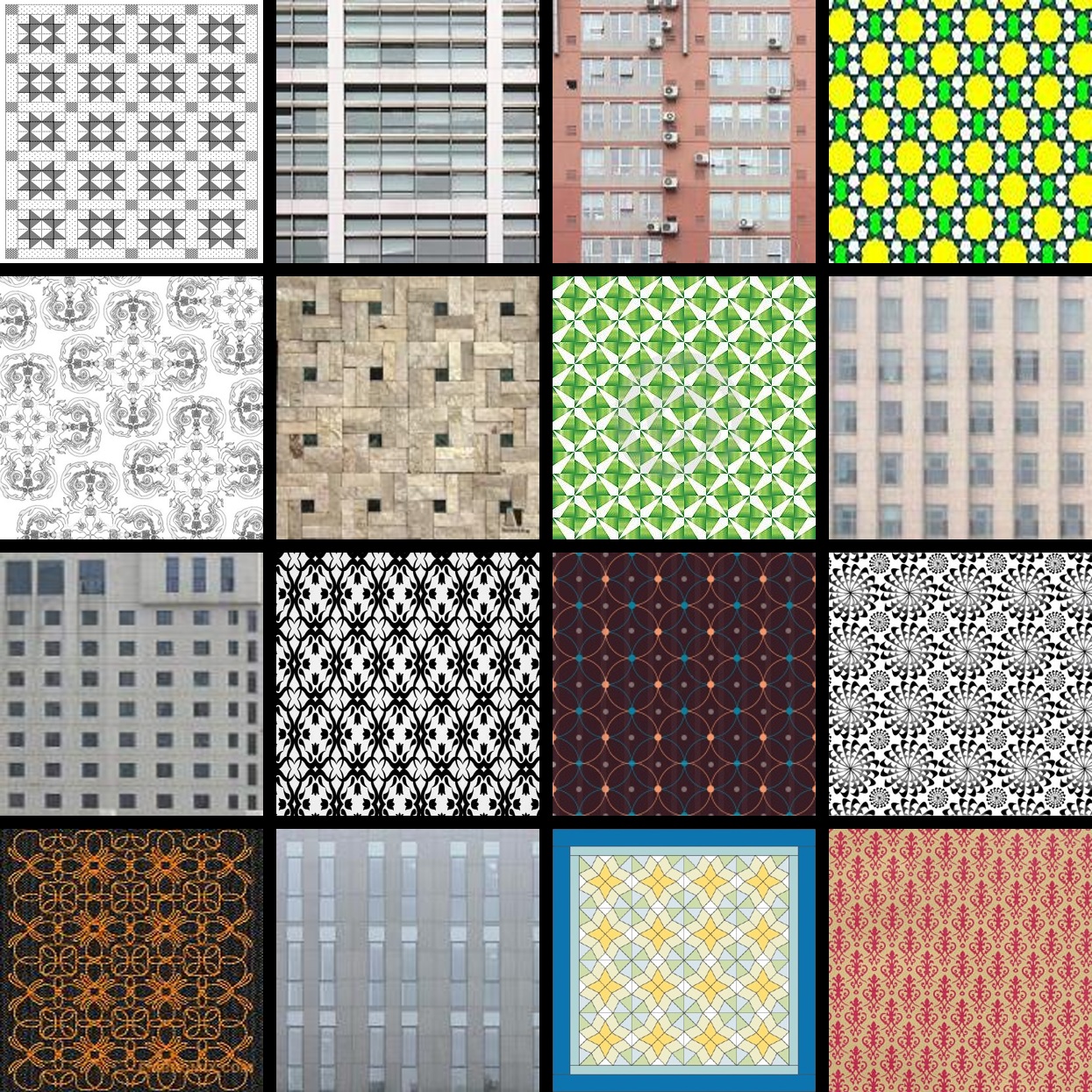}\hspace{5mm}
\includegraphics[width=1.05\columnwidth]{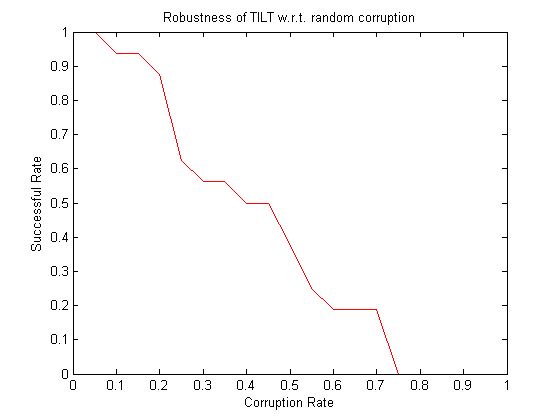}
}
\caption{{\bf Robustness Tests of TILT} on various low-rank textures. The textures on the left are ordered in descent order of being robust to random corruption: from left to right, from top to bottom. Plot on the right shows TILT succeeds with how many textures at each level of corruption.}
\label{fig:robustness2}
\end{figure*}

In this experiment, we test the robustness of TILT on some representative synthetic and realistic low-rank patterns, shown in Figure \ref{fig:robustness2} (left). We introduce a small deformation to each texture (say rotation by $10^\circ$) and examine if TILT converges to the correct solution under different levels of random corruption. We randomly select a fraction (from 0\% to 100\%) of the pixels and assign them a random value in the range (0, 255). We run the TILT algorithm on such corrupted images and examine how many images are correctly rectified by TILT at each level of corruption. The results are shown in Figure \ref{fig:robustness2} (right).
Notice that for almost three quarters of the textures TILT can tolerate up to 30\% random corruption and for textures in the first row TILT can rectify the deformation even if more than 50\% of the pixels are corrupted. Also, we notice that for textures where TILT has low error tolerance, their textures either have very low contrast, or are rather sparse, or have relatively high rank.

\begin{figure*}[!ht]
\centerline
{
    \subfigure
    {
        \includegraphics[width=0.21\textwidth]{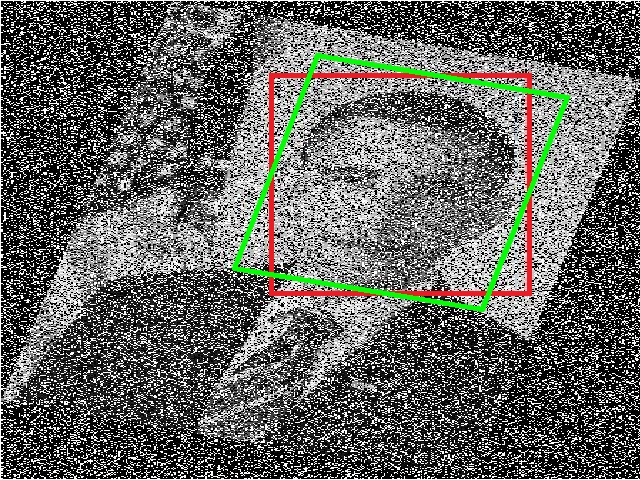}
    }
    \subfigure
    {
        \includegraphics[width=0.21\textwidth]{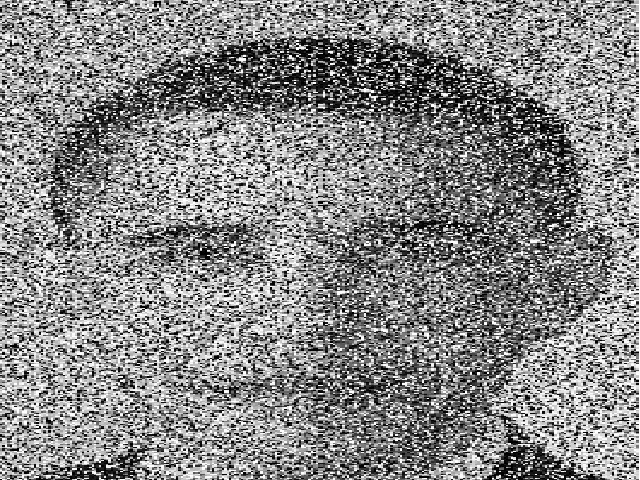}
    }
    \subfigure
    {
        \includegraphics[width=0.21\textwidth]{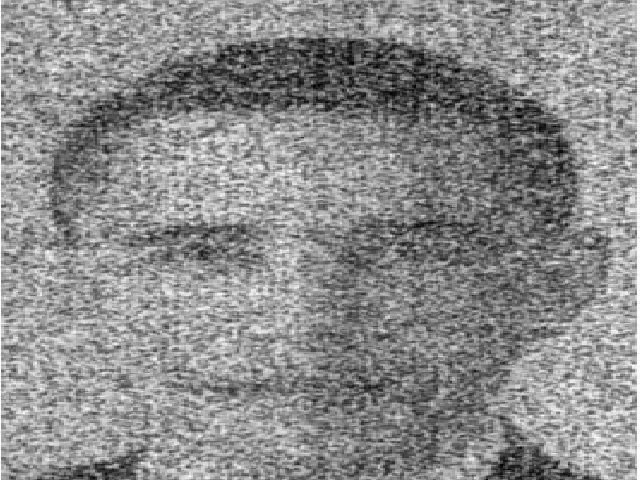}
    }
    \subfigure
    {
        \includegraphics[width=0.21\textwidth]{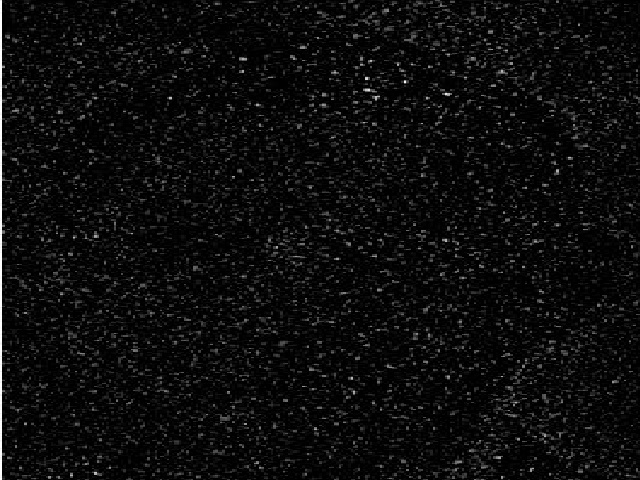}
    }
}
\centerline
{
    \subfigure
    {
        \includegraphics[width=0.21\textwidth]{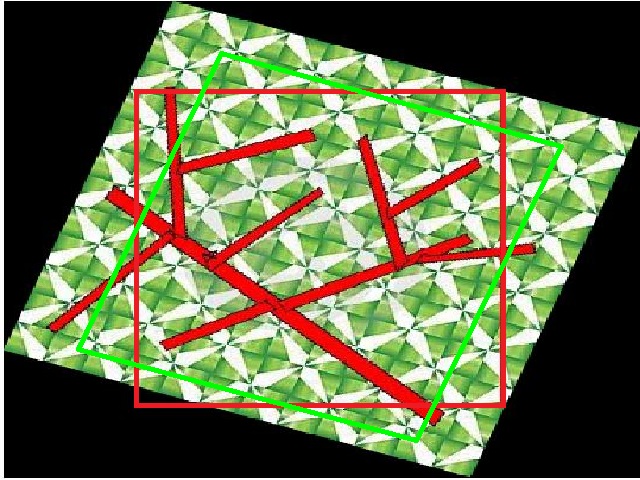}
    }
    \subfigure
    {
        \includegraphics[width=0.21\textwidth]{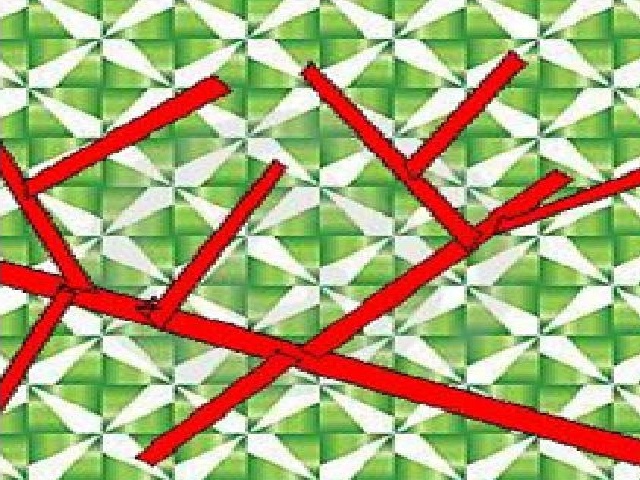}
    }
    \subfigure
    {
        \includegraphics[width=0.21\textwidth]{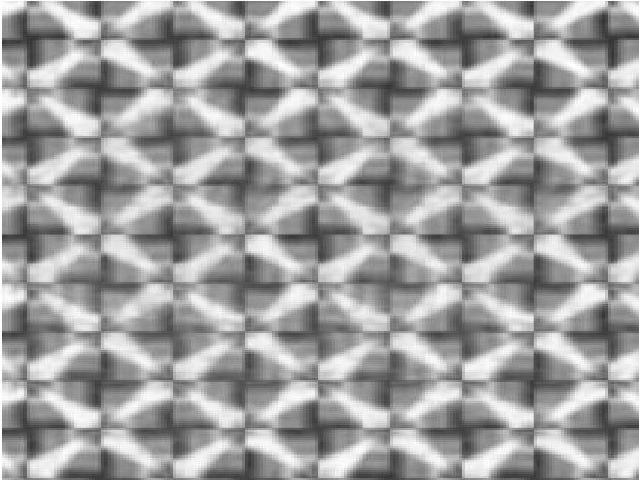}
    }
    \subfigure
    {
        \includegraphics[width=0.21\textwidth]{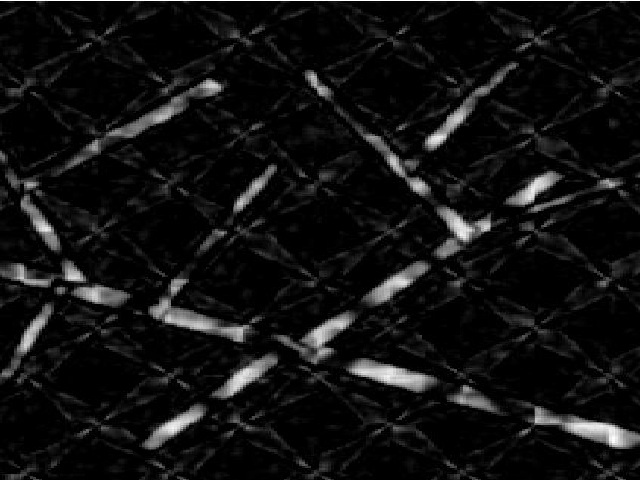}
    }
}
\centerline
{
    \subfigure[Input $I$]
    {
        \includegraphics[width=0.21\textwidth]{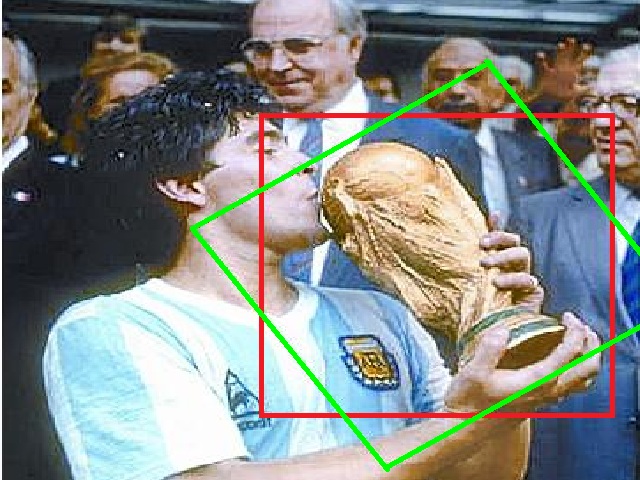}
    }
    \subfigure[Output $I\circ \tau$]
    {
        \includegraphics[width=0.21\textwidth]{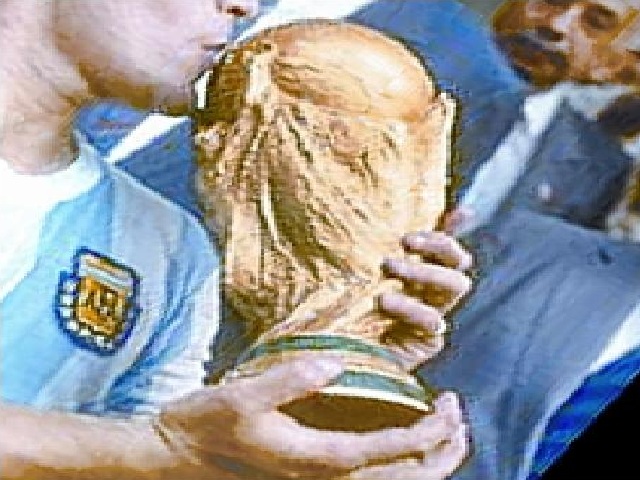}
    }
    \subfigure[Low rank $I^0$]
    {
        \includegraphics[width=0.21\textwidth]{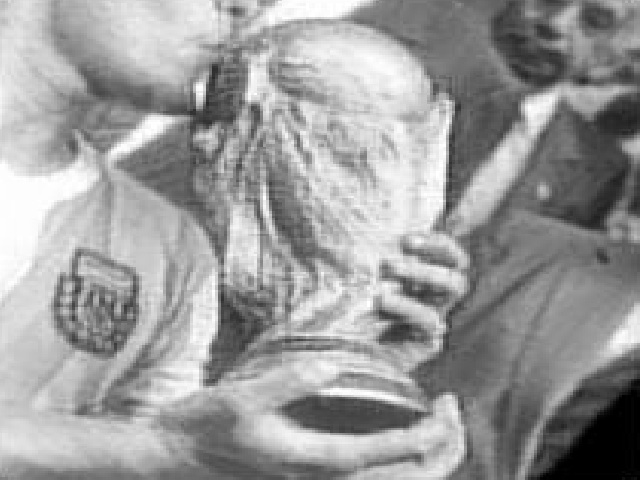}
    }
    \subfigure[Sparse error $E$]
    {
        \includegraphics[width=0.21\textwidth]{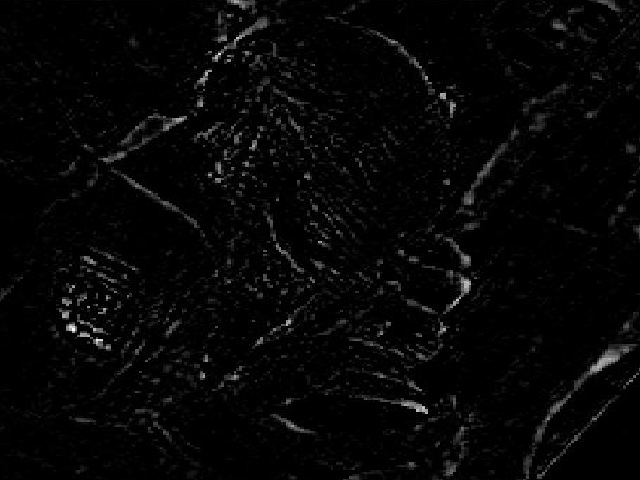}
    }
}
\caption{{\bf Robustness of TILT.} Top row: random corruption added to 60\% pixels; Middle row: scratches added on a symmetric pattern; Bottom row: containing cluttered background.}
\label{fig:robustness}
\end{figure*}

Figure \ref{fig:robustness} show some more examples for the robustness of the proposed algorithm to random corruption, occlusions, and cluttered background, respectively. For the first two experiments, no branch-and-bound is even needed.

The above experiments demonstrate the robustness of TILT to randomly located corruptions. However, in some cases we may have some idea about which part of the images are likely to be corrupted or occluded. For instance, if the initial window is too close to the image boundary, the algorithm may converge to a region outside of the image boundary. In such cases, we know which pixels in the region are missing. This information can help us to modify the algorithm and further improve its robustness. We will discuss this case in more detail in Section \ref{sec:extensions} when we study possible extensions to TILT.

\subsection{Shape from Low-rank Textures}
\begin{figure*}[!ht]
\centerline{
\includegraphics[width=0.43\textwidth]{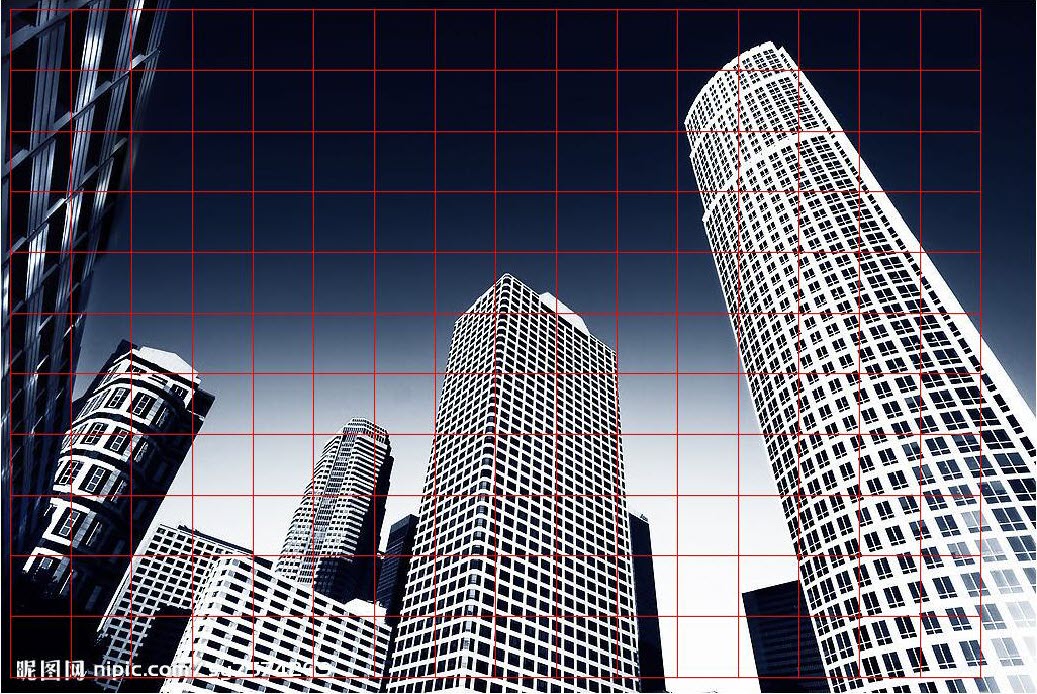} \hspace{6mm}
\includegraphics[width=0.43\textwidth]{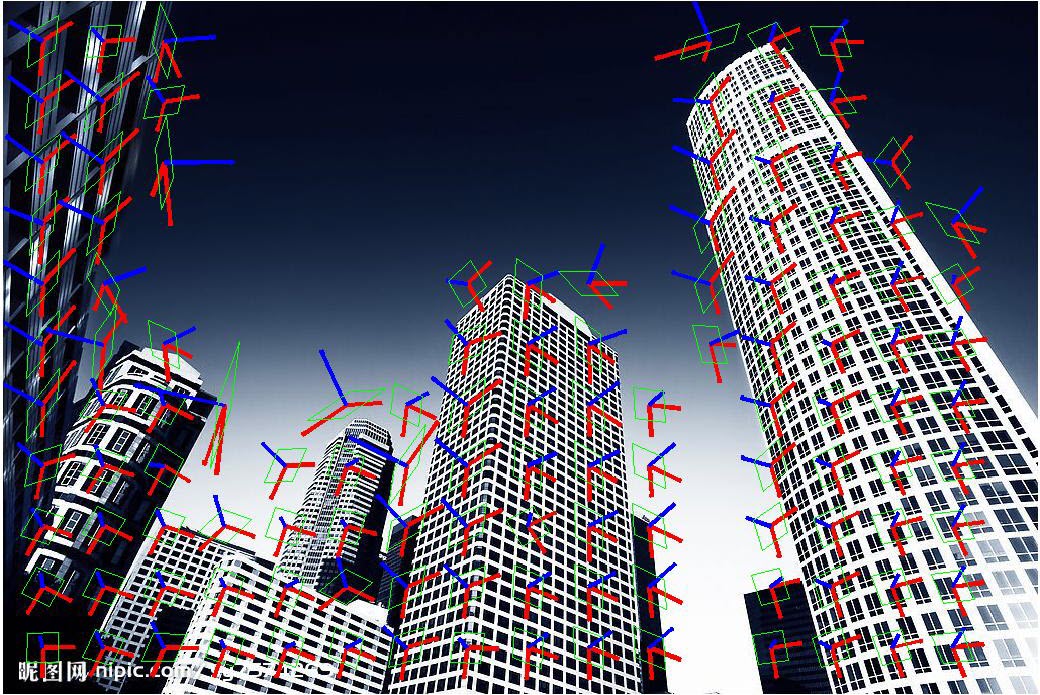}
}
\centerline{
\includegraphics[width=0.43\textwidth]{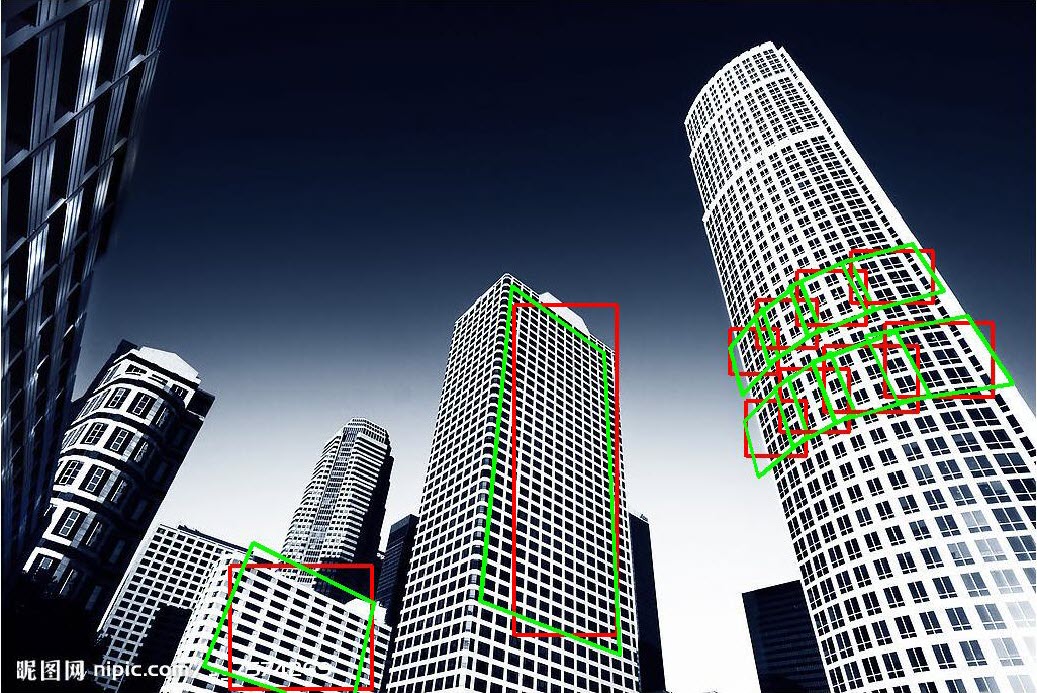} \hspace{6mm}
\includegraphics[width=0.43\textwidth]{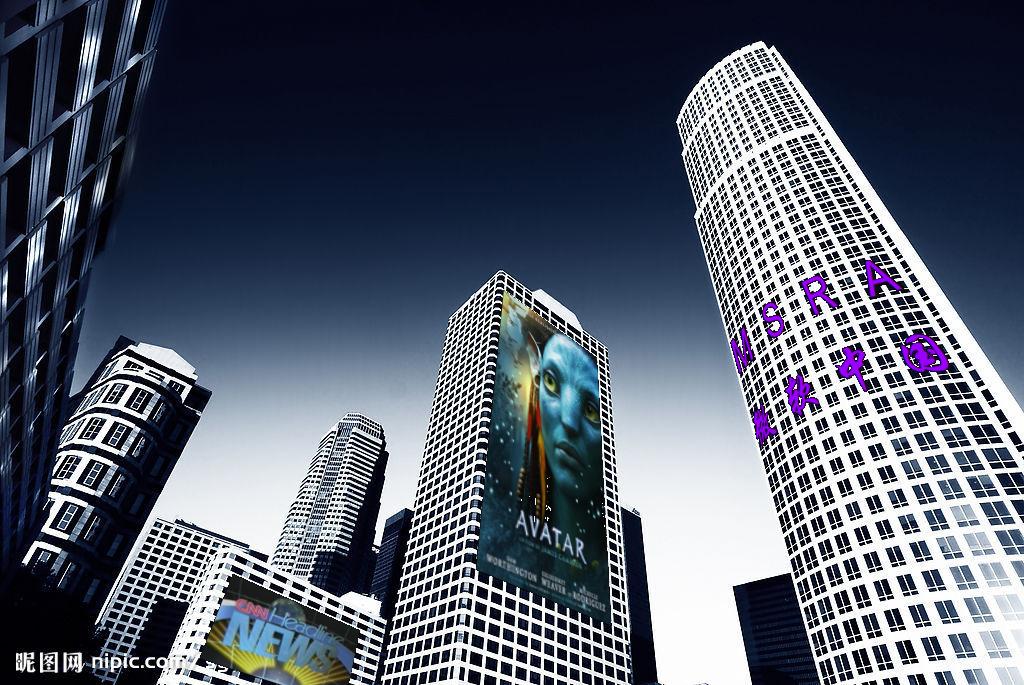}
}
\centerline{
\includegraphics[width=0.25\textwidth]{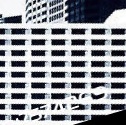} \hspace{2mm}
\includegraphics[width=0.08\textwidth]{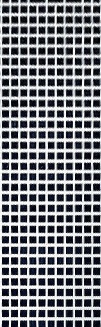}\hspace{2mm}
\includegraphics[height=0.13\textwidth]{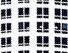}
\includegraphics[height=0.13\textwidth]{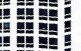}
\includegraphics[height=0.13\textwidth]{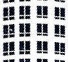}
}
\caption{{\bf Shape from (Low-rank) Textures.} Top left: The input grid of $60\times 60$ windows. Top right: Low-rank textures detected by the TILT algorithm with affine transform and the recovered local affine geometry. Middle left: Use homography to get the projective transformations. Middle right: the resulting image with the marked regions augmented with virtual objects. Bottom row: representative low-rank textures recovered from the marked regions of the buildings.}
\label{fig:geometry}
\end{figure*}

Obviously, the rectified low-rank textures found by our algorithm can facilitate many vision tasks, including establishing correspondences among images, recognizing text and objects, or reconstructing the 3D structure of a scene, etc. Due to limited space, we only illustrate how our algorithm can help extract precise, rich geometric and structural information from an image of an urban scene, as shown in Figure \ref{fig:geometry} (top). This complements many existing ``Shape from X'' methods in the vision literature.

The size of the image shown in Figure \ref{fig:geometry} is $1024\times 685$ pixels and we simply run the TILT algorithm (with affine transforms) on a grid of $60\times 60$ windows. If the rank of the resulting texture drops significantly from that of the original window, we say that the algorithm has ``detected" a region with some low-rank structure.\footnote{The image rank is computed by thresholding the singular values at 1/30th of the largest one. Also, we throw away regions whose largest singular value is too small, which typically correspond to a smooth region like the sky.} In Figure \ref{fig:geometry}, we have shown the resulting deformed windows, together with the local orientation and surface normal recovered from the recovered affine transformation. Notice that for windows located inside the building facades, TILT correctly recovers the local geometry for almost all of them; even for patches located at the edge of the facades, one of the sides of the rectified patches always aligns precisely with the building's edge.

Of course, one can initialize the size of the windows at different sizes or scales. But for larger regions, affine transforms will not be accurate enough to describe the deformation caused by a perspective projection. For instance, the entire facade of the middle building in Figure \ref{fig:geometry} (middle row) obviously exhibits significant projective deformation. Nevertheless, if we initialize the projective TILT algorithm with the output from the affine TILT algorithm on a small patch on the facade, the algorithm can easily converge to the correct homography and recover the low-rank textures correctly, as shown in Figure \ref{fig:geometry} (middle row).

With both the low-rank texture and their geometry correctly recovered, we can easily perform many interesting tasks such as editing parts of the images while respecting the true 3D shape and the correct perspective. Figure \ref{fig:geometry} (middle row) shows some examples, which suggest that our method can be very useful for many augmented-reality related applications.

\subsection{Rectifying Many Categories of Low-rank Textures}
In this section, we test the efficacy of the TILT algorithm on natural images belonging to various categories. Besides some examples where TILT works very well, we also present some cases that are particularly challenging where our algorithm succeeds only to some extent, and some examples where it fails. We believe that from these examples, the readers may gain a better understanding about both the strength and limitations of the TILT algorithm.

\begin{figure*}[!ht]
\centerline{
\subfigure{
\includegraphics[width=0.22\textwidth]{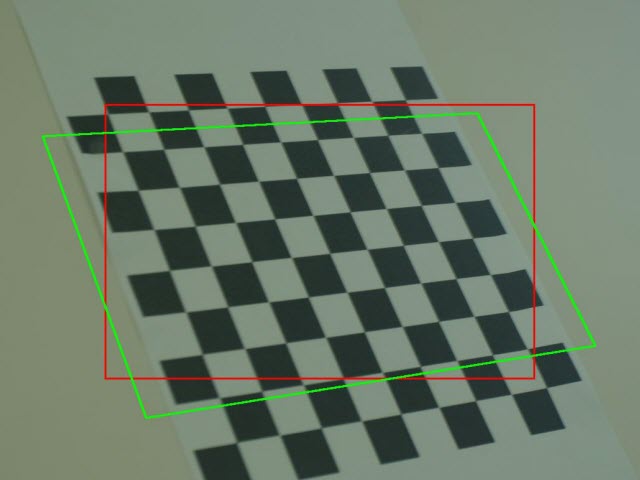}
}
\subfigure{
\includegraphics[width=0.22\textwidth]{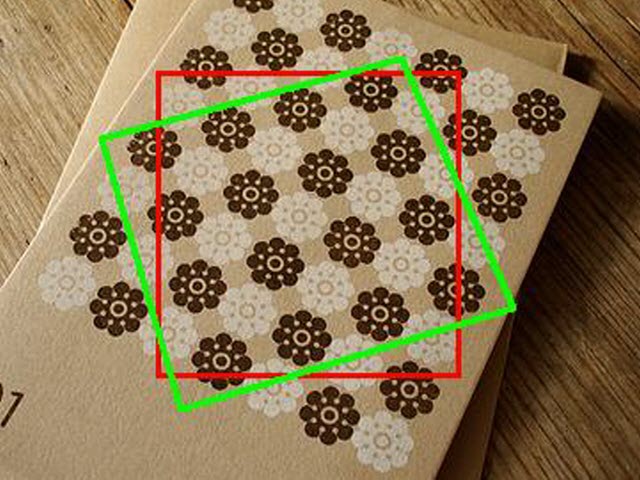}
}
\subfigure{
\includegraphics[width=0.22\textwidth]{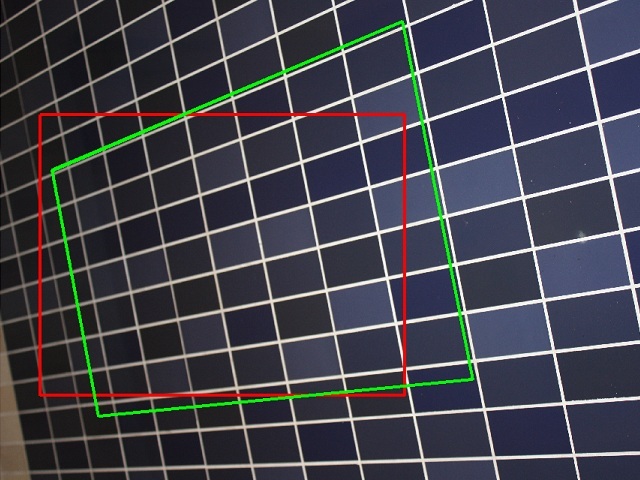}
}
\subfigure{
\includegraphics[width=0.22\textwidth]{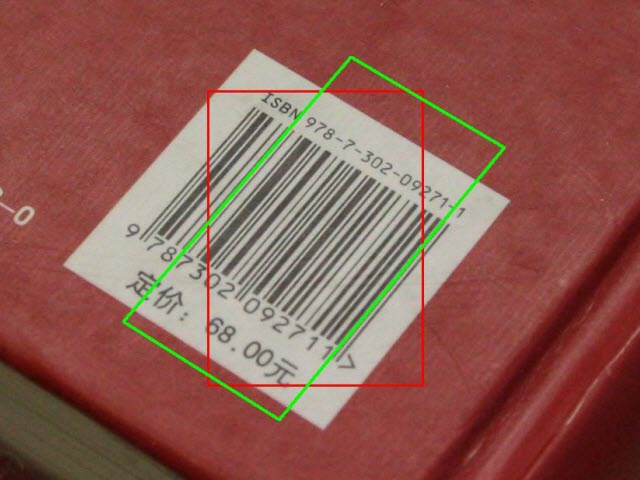}
}
}

\centerline{
\subfigure{
\includegraphics[width=0.22\textwidth]{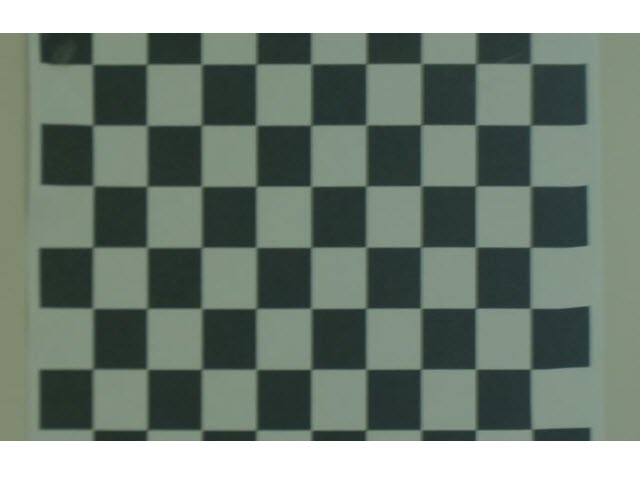}
}
\subfigure{
\includegraphics[width=0.22\textwidth]{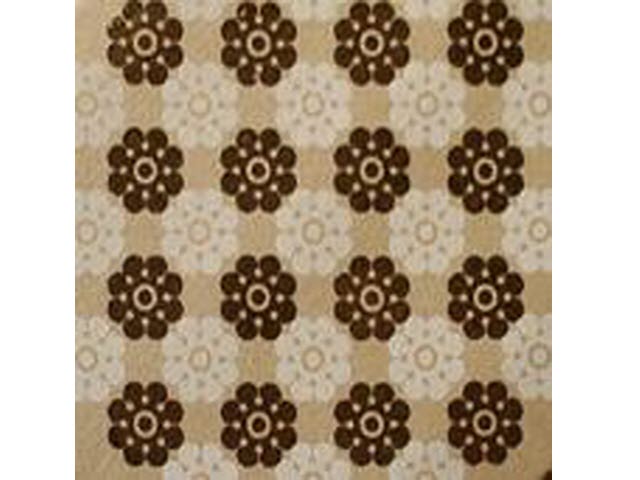}
}
\subfigure{
\includegraphics[width=0.22\textwidth]{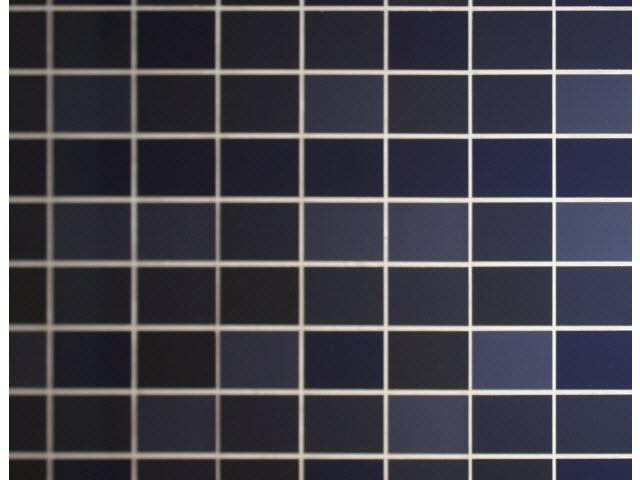}
}
\subfigure{
\includegraphics[width=0.22\textwidth]{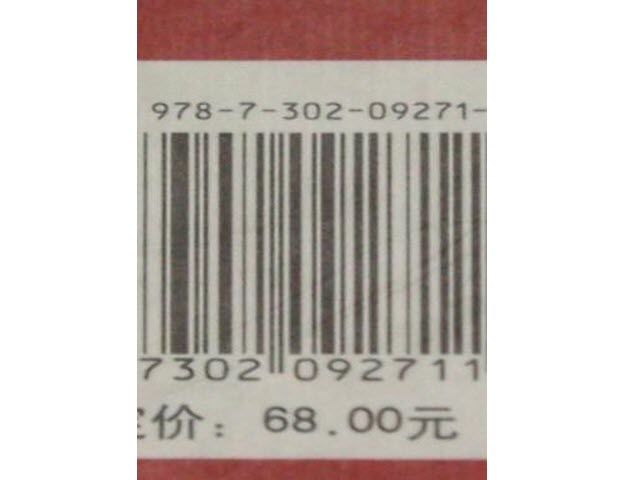}
}
}

\centerline{
\subfigure{
\includegraphics[width=0.22\textwidth]{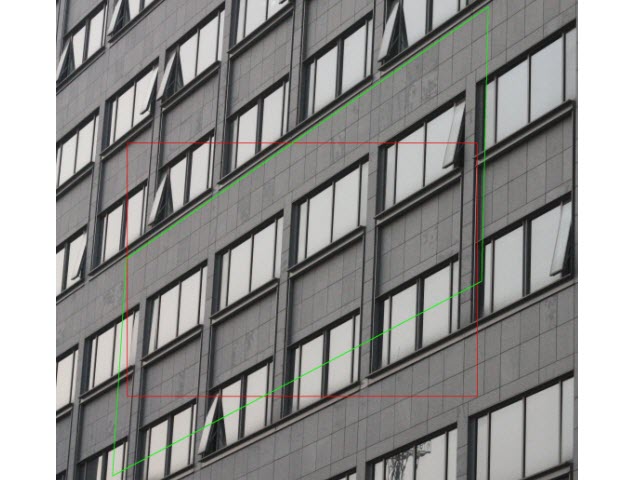}
}
\subfigure{
\includegraphics[width=0.22\textwidth]{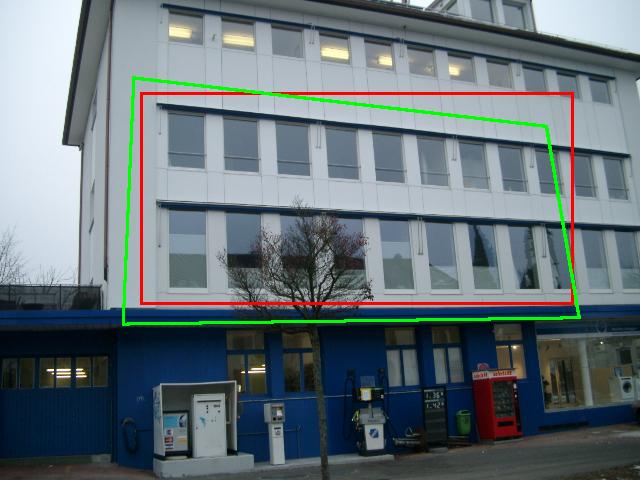}
}
\subfigure{
\includegraphics[width=0.22\textwidth]{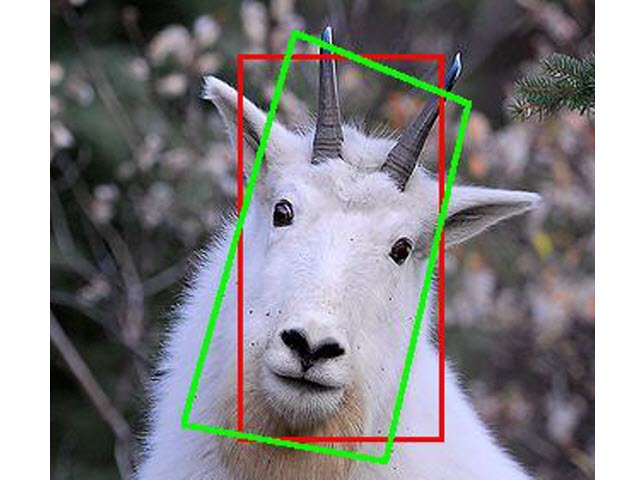}
}
\subfigure{
\includegraphics[width=0.22\textwidth]{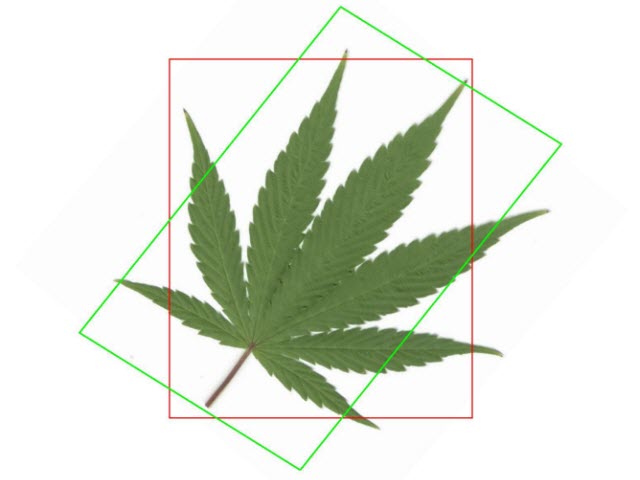}
}
}

\centerline{
\subfigure{
\includegraphics[width=0.22\textwidth]{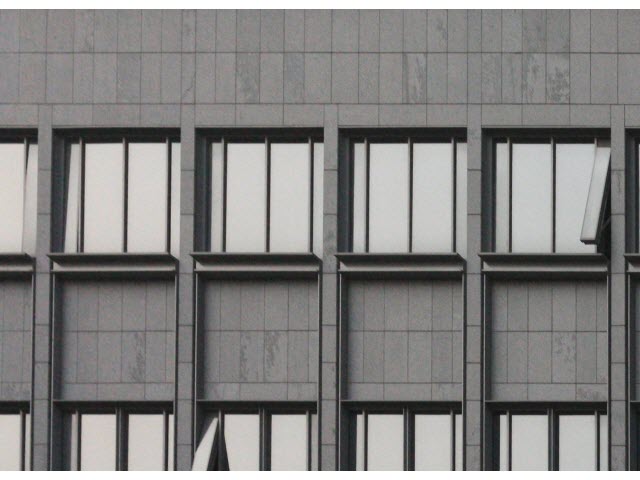}
}
\subfigure{
\includegraphics[width=0.22\textwidth]{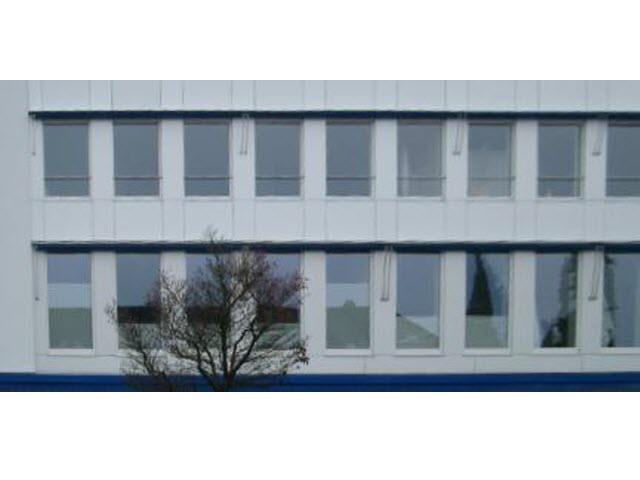}
}
\subfigure{
\includegraphics[width=0.22\textwidth]{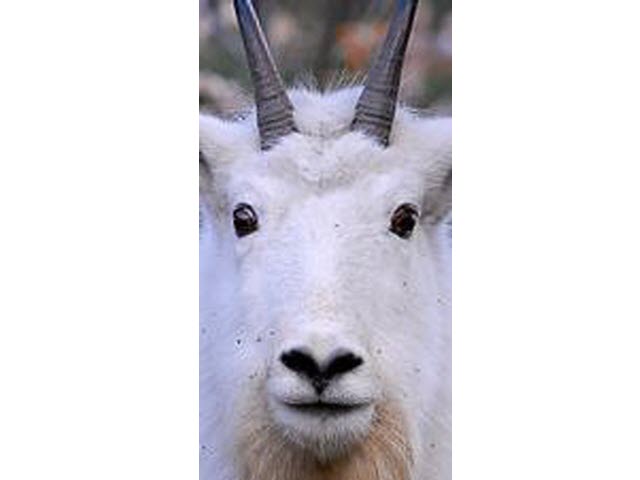}
}
\subfigure{
\includegraphics[width=0.22\textwidth]{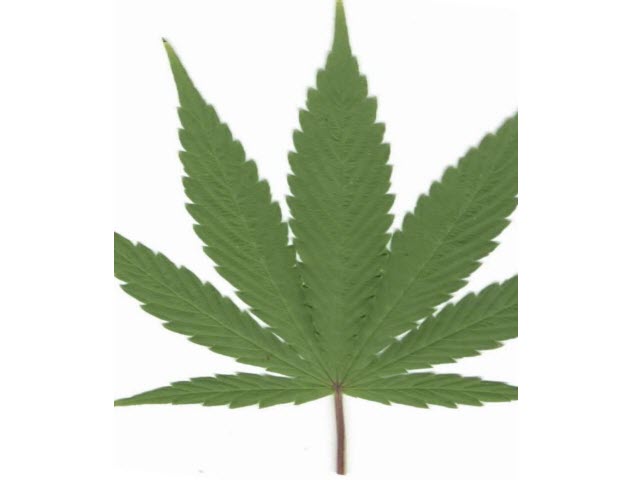}
}
}\centerline{
\subfigure{
\includegraphics[width=0.22\textwidth]{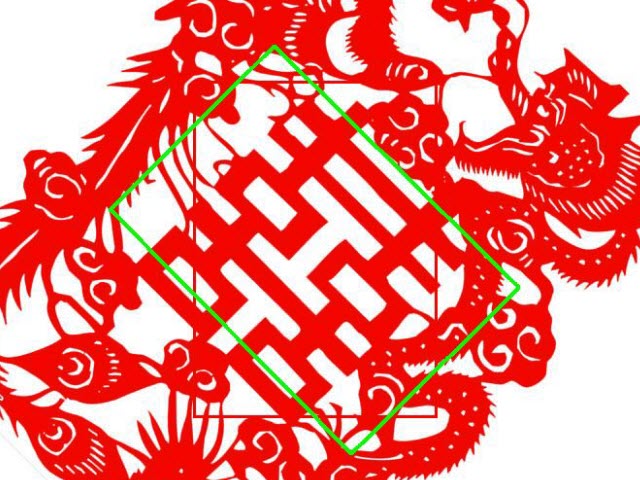}
}
\subfigure{
\includegraphics[width=0.22\textwidth]{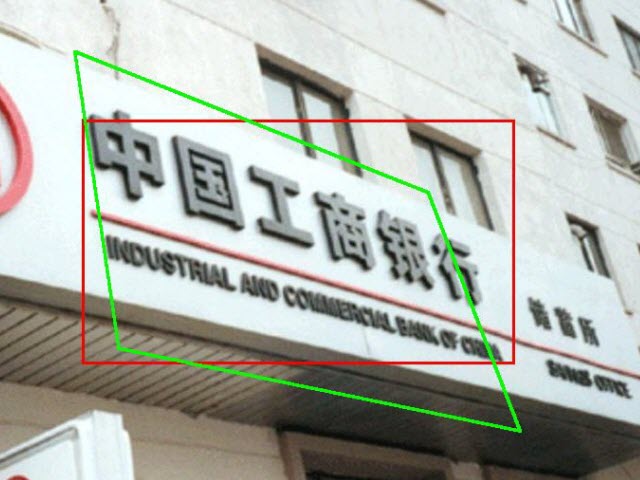}
}
\subfigure{
\includegraphics[width=0.22\textwidth]{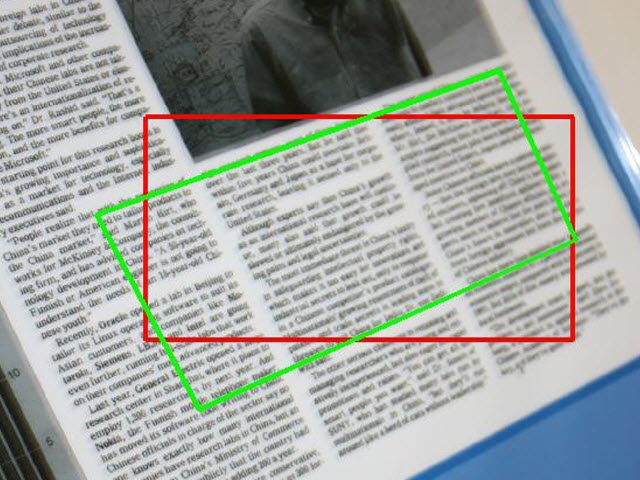}
}
\subfigure{
\includegraphics[width=0.22\textwidth]{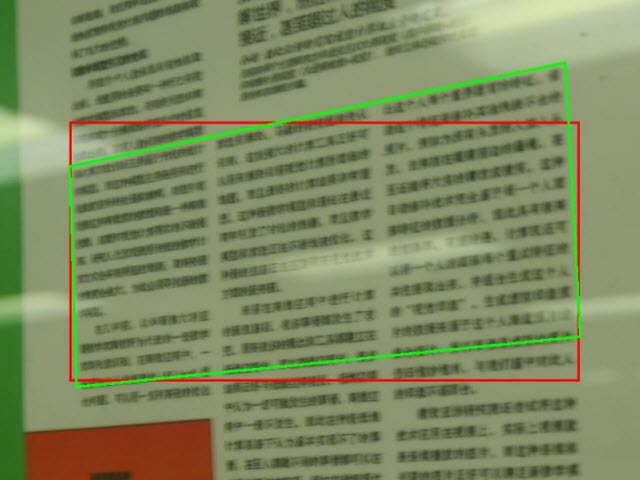}
}
}

\centerline{
\subfigure{
\includegraphics[width=0.22\textwidth]{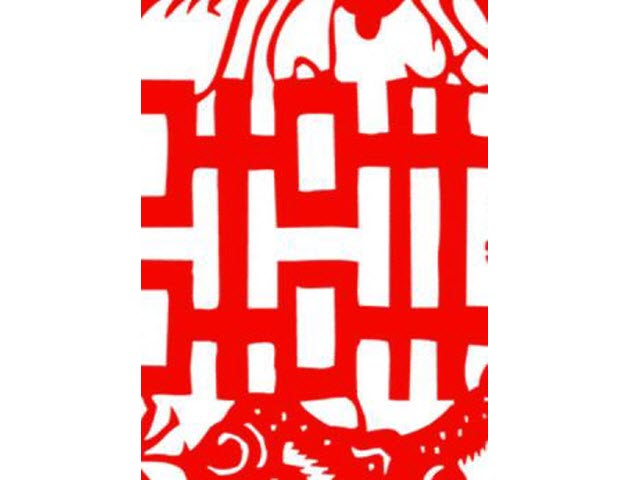}
}
\subfigure{
\includegraphics[width=0.22\textwidth]{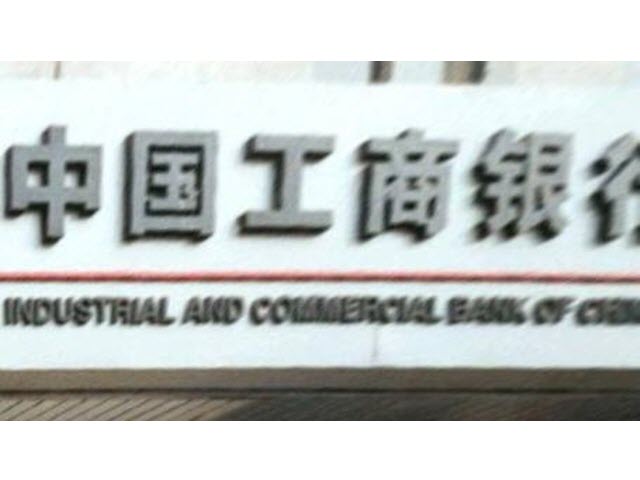}
}
\subfigure{
\includegraphics[width=0.22\textwidth]{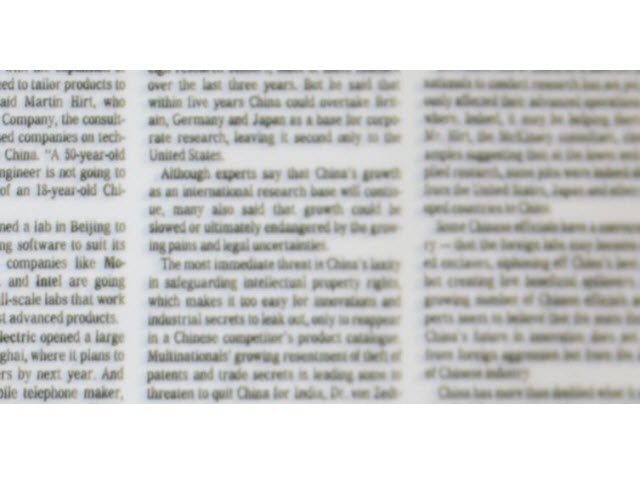}
}
\subfigure{
\includegraphics[width=0.22\textwidth]{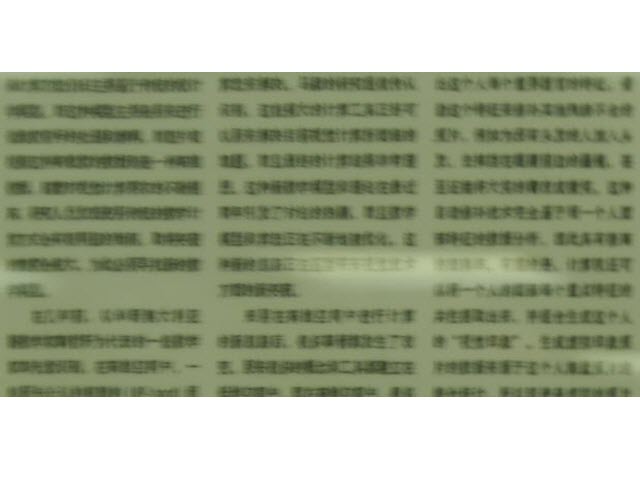}
}
}
\caption{{\bf Representative Results of TILT.} The objects can be categorized as follows. Top two rows: regular patterns and textures; Middle two rows: signs, characters, and printed text; Bottom two rows: bar code, objects with bilateral symmetry. In each case, the red window denotes the input and the green window denotes the final output. The image enclosed by the green window is rectified and displayed to emphasize the low-rank structure.}
\label{fig:categories}
\end{figure*}

Since the proposed TILT algorithm has a decent range of convergence for both affine and projective deformations and it is also very robust to sparse corruption of the image intensity, we find that it works remarkably well for a very broad range of patterns, regular structures, natural objects and even printed text with an approximate low-rank structure. Figure \ref{fig:categories} shows many such examples, from which we see that even when initialized with a very rough rectangular window, our algorithm can converge precisely to the underlying low-rank structure of the images, despite occlusion, noisy background, illumination change, and significant deformation.

\begin{figure*}[!ht]
\centerline{
\subfigure[boundary effect]{
\includegraphics[width=0.25\textwidth]{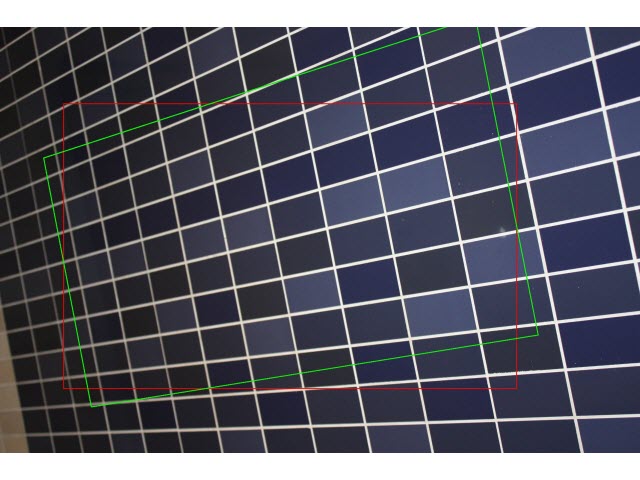}
}
\subfigure[lack of regularity]{
\includegraphics[width=0.25\textwidth]{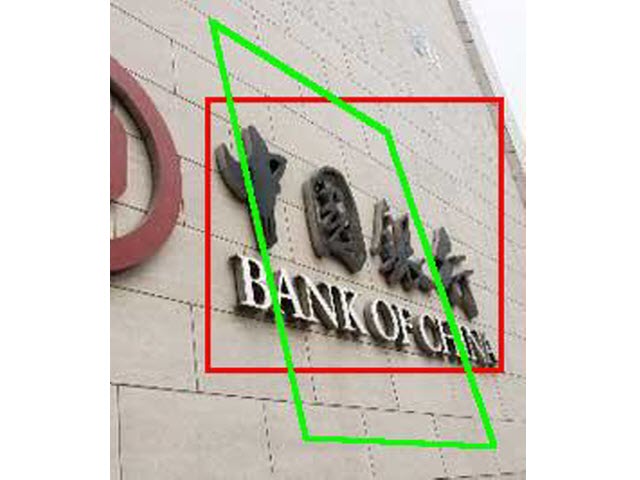}
}
\subfigure[non-planar]{
\includegraphics[width=0.25\textwidth]{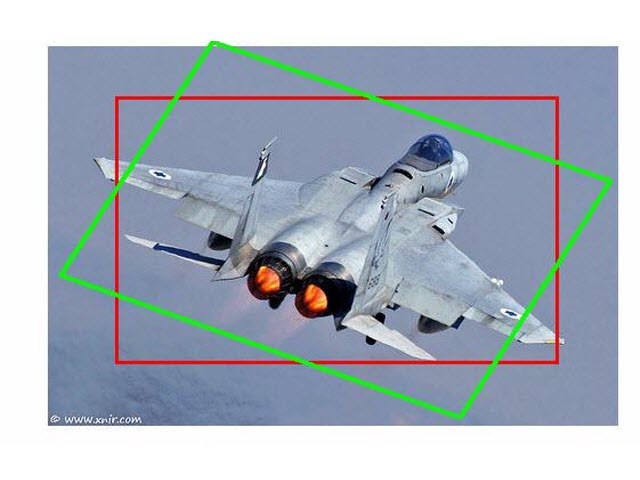}
}
}
\centerline{
\subfigure{
\includegraphics[width=0.25\textwidth]{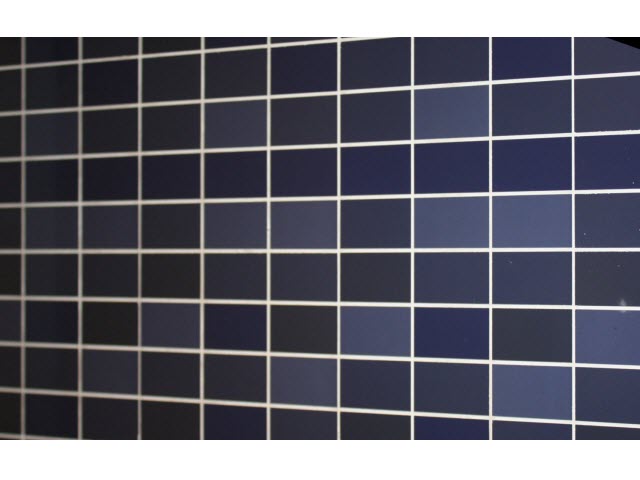}
}
\subfigure{
\includegraphics[width=0.25\textwidth]{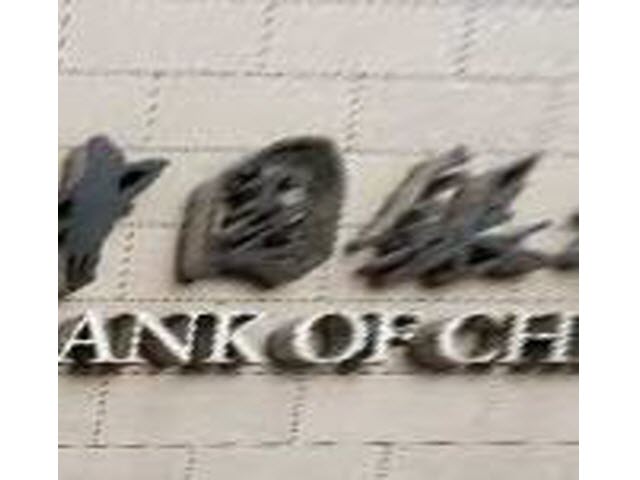}
}
\subfigure{
\includegraphics[width=0.25\textwidth]{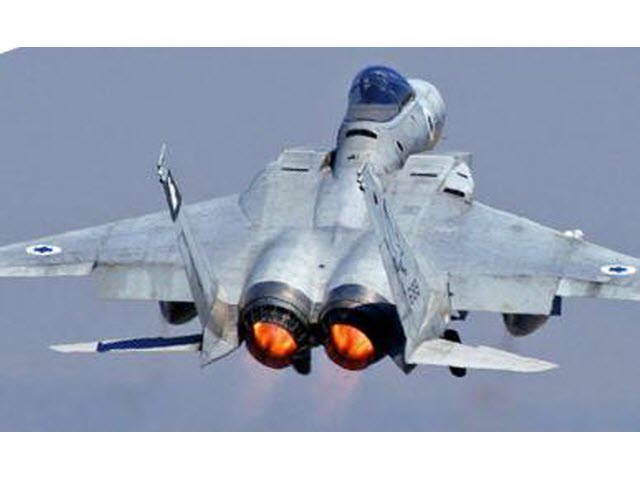}
}
}
\centerline{
\subfigure[large defomration]{
\includegraphics[width=0.25\textwidth]{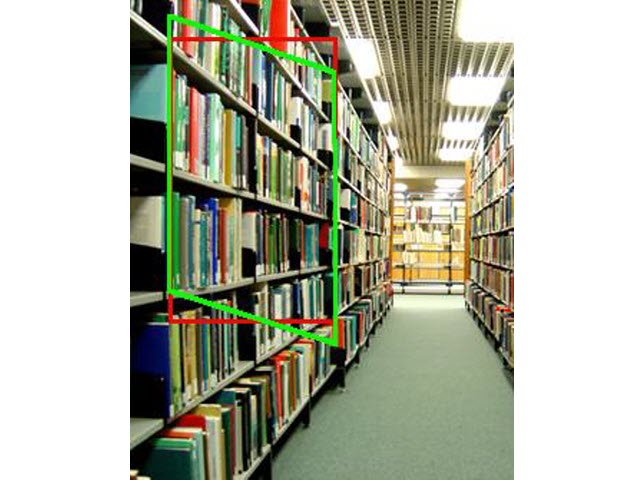}
}
\subfigure[too much background]{
\includegraphics[width=0.25\textwidth]{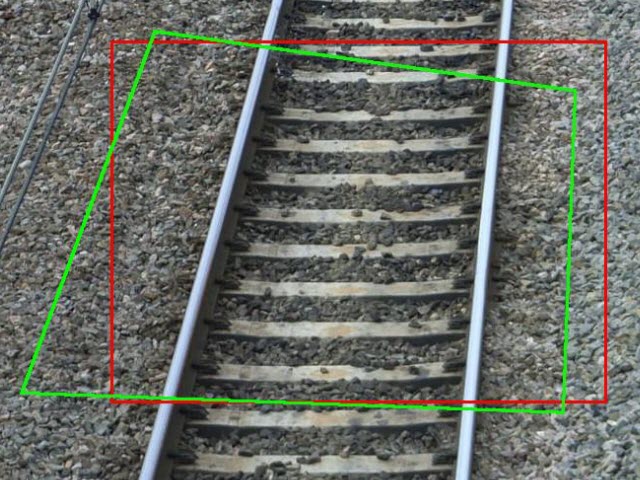}
}
\subfigure[sparse regular structures]{
\includegraphics[width=0.25\textwidth]{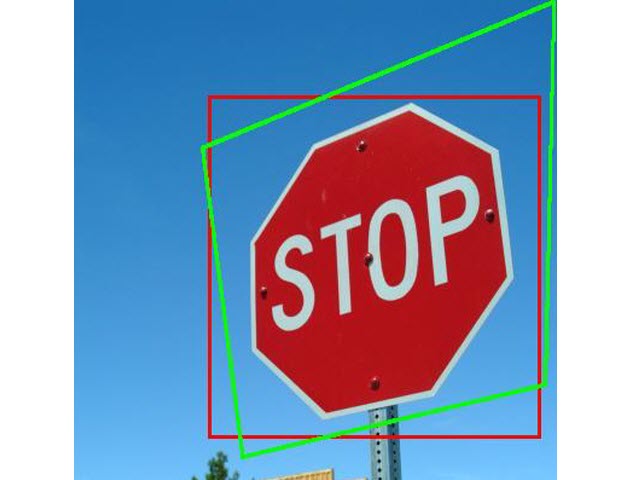}
}
}
\centerline{
\subfigure{
\includegraphics[width=0.25\textwidth]{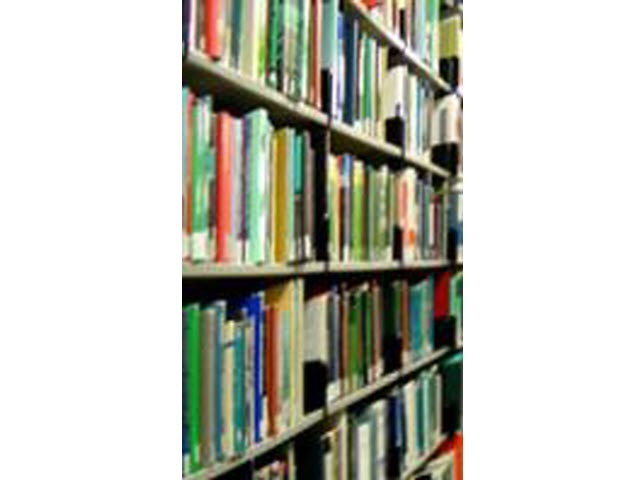}
}
\subfigure{
\includegraphics[width=0.25\textwidth]{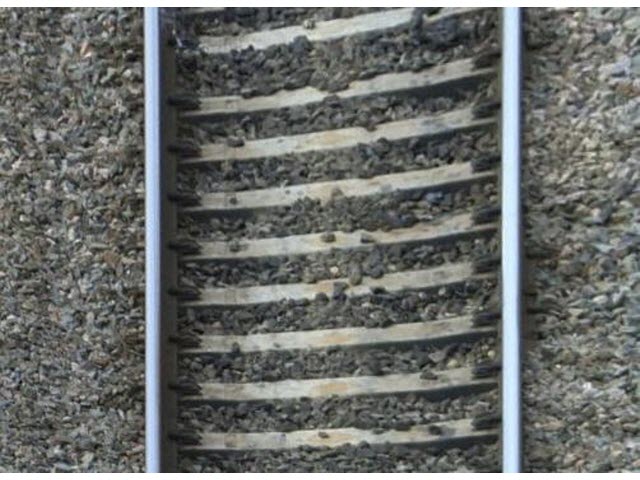}
}
\subfigure{
\includegraphics[width=0.25\textwidth]{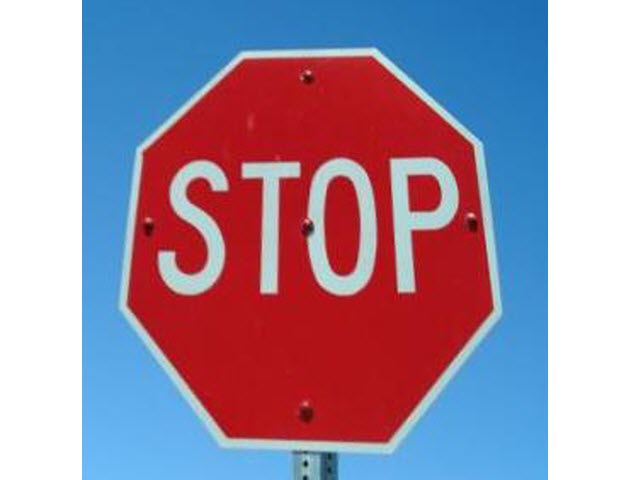}
}
}
\caption{{\bf Challenging Cases.} TILT converges to an approximately correct solution at best for these examples. Top: from left to right: boundary problem, not enough regular texture, non-planar objects. Bottom: from left to right: large perspective distortion; too much random texture in background; sparse (binary image) low-rank structure.}
\label{fig:challenging-cases}
\end{figure*}

\begin{figure}
\centerline{
\includegraphics[width=0.25\textwidth]{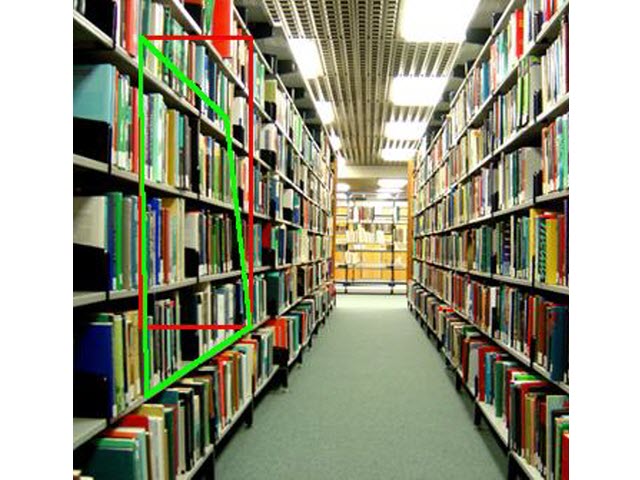}\includegraphics[width=0.25\textwidth]{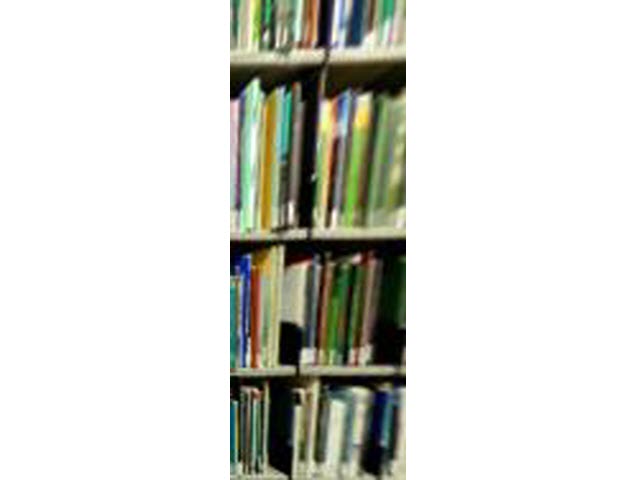}
}
\centerline{
\includegraphics[width=0.22\textwidth]{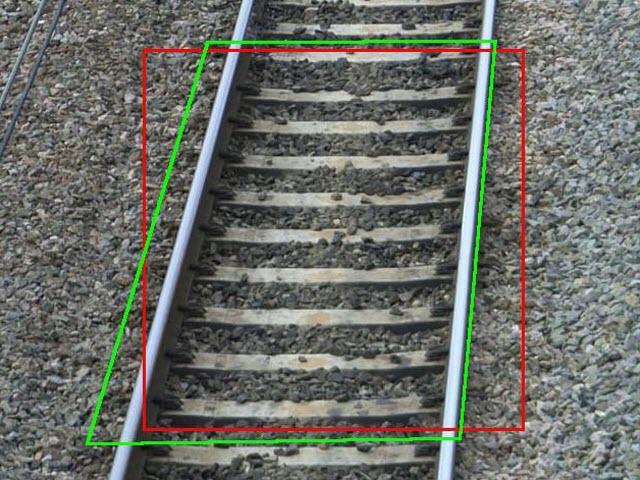}
\includegraphics[width=0.22\textwidth]{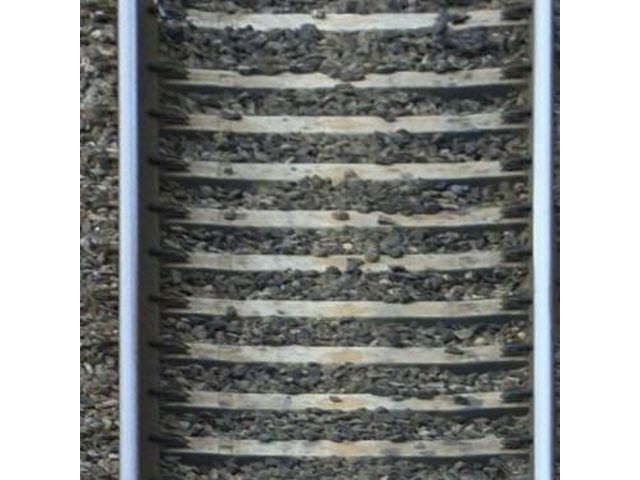}
}
\caption{{\bf Effect of Initialization.} For the examples in Figure \ref{fig:challenging-cases}(g) and (h) where TILT had failed earlier, the correct solution is found with a slightly different initialization, in both cases by reducing the horizontal width of the initial (red) window.}
\label{fig:large-tilt}
\end{figure}

\paragraph{Issues with more challenging cases.} One should expect our algorithm to work well only when the low-rank and sparse structure assumptions, explained in Section \ref{sec:formulation}, hold true. The current algorithm is only a basic version and its capability is still limited, especially when we try to apply it to cases where the assumptions are not fully met. Through the remainder of this section and the next section, we will discuss some of the limitations of TILT, as well as potential extensions that make it work better in some of the more challenging cases. Figure \ref{fig:challenging-cases} shows some examples on which TILT does not perform as well as it did in previous examples. These examples are arguably more challenging than those shown in Figure \ref{fig:categories}:
\begin{itemize}
\item Figure \ref{fig:challenging-cases}(a) is an example when the size of the input window is too large. Ideally, the correct solution is supposed to converge to a region beyond the image boundary. It stops once it hits the boundary which is a only  partially correct solution. In the next section, we will show how this problem can be addressed by combining the basic TILT algorithm with techniques from low-rank matrix completion.
\item Figure \ref{fig:challenging-cases}(b) shows a case where the algorithm manages to converge to an approximate solution despite the fact that there is a lack of regularity in the printed text. TILT managed to correct the perspective distortion partially in this case.
\item Figure \ref{fig:challenging-cases}(c) shows a case where the algorithm manages to correct the overall pose of the object despite fact that the object is not planar, similar to some of the cases shown earlier in Figure \ref{fig:TILT-overview}.
\item Figure \ref{fig:challenging-cases}(g) shows a failed case, where the perspective deformation is too large for the given input window and the texture is complex (the rank is relatively high). Nevertheless, with slightly better initialization,\footnote{say by aggregating TILT results from smaller affine patches or using rough manual inputs} we expect the TILT algorithm to converge to the correct solution. For example, as shown in Figure \ref{fig:large-tilt} (top), if we simply shorten the width of the initial window along the main tilted direction, the algorithm manages to find the correct solution.
\item Figure \ref{fig:challenging-cases}(h) shows another failed case, where the initial window contains too much of the background, which has the appearance of a random texture with little structure, the algorithm converges to a local minimum. Nevertheless, with a slightly different initial window that contains less background, the algorithm converges to the correct solution (see Figure \ref{fig:large-tilt} (bottom)).
\item Figure \ref{fig:challenging-cases}(i) shows a case where the low-rank texture itself is close to a sparse binary image. The algorithm only manages to converge to a partially correct transform -- the recovered texture is approximately symmetric along the horizontal direction. In this case, in order to improve the results, we may have to adjust the weights between the low-rank and sparse components in the cost function in \eqref{eqn:convex_linear_tilt}, or to enforce the symmetry of the desired solution explicitly in the form of additional constraints.
\end{itemize}

\begin{figure*}[!ht]
\centerline{
\subfigure[high-rank structures]{
\includegraphics[width=0.22\textwidth]{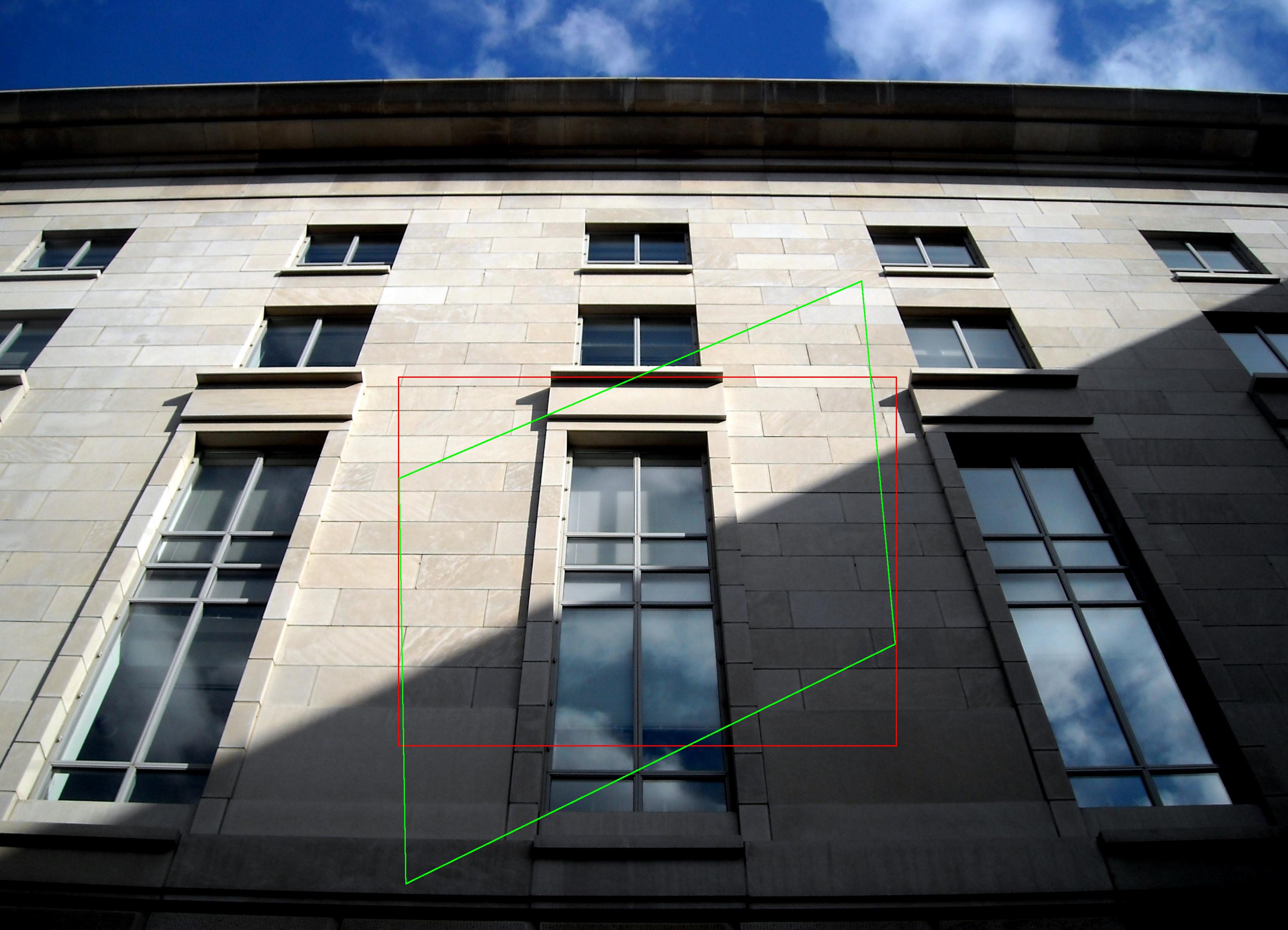}
}
\subfigure[two low-rank regions]{
\includegraphics[width=0.22\textwidth]{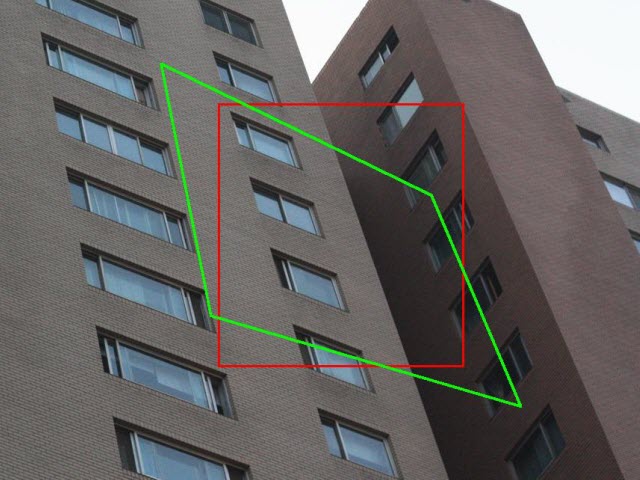}
}
\subfigure[too much occlusion]{
\includegraphics[width=0.24\textwidth]{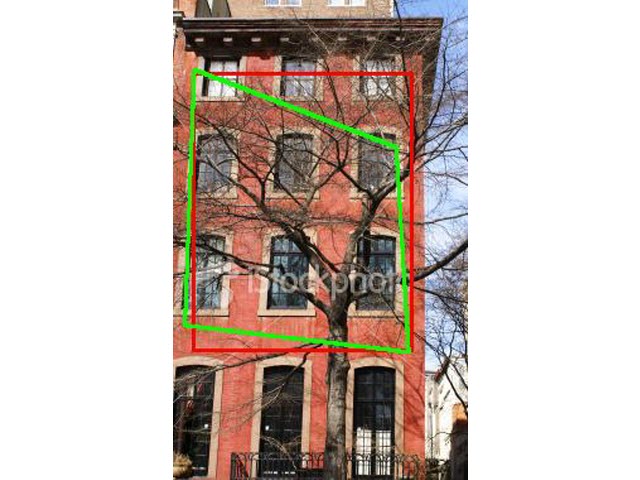}
}
\subfigure[random textures]{
\includegraphics[width=0.22\textwidth]{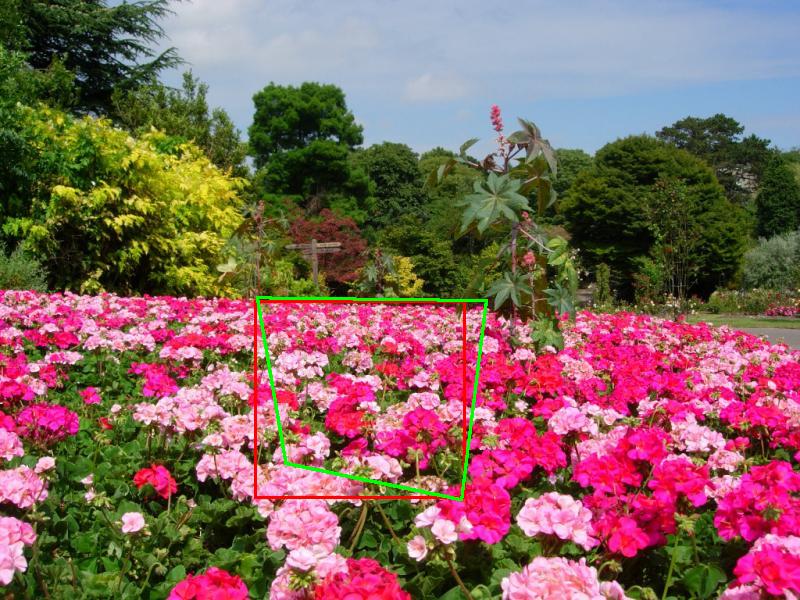}
}
}
\centerline{
\subfigure{
\includegraphics[width=0.22\textwidth]{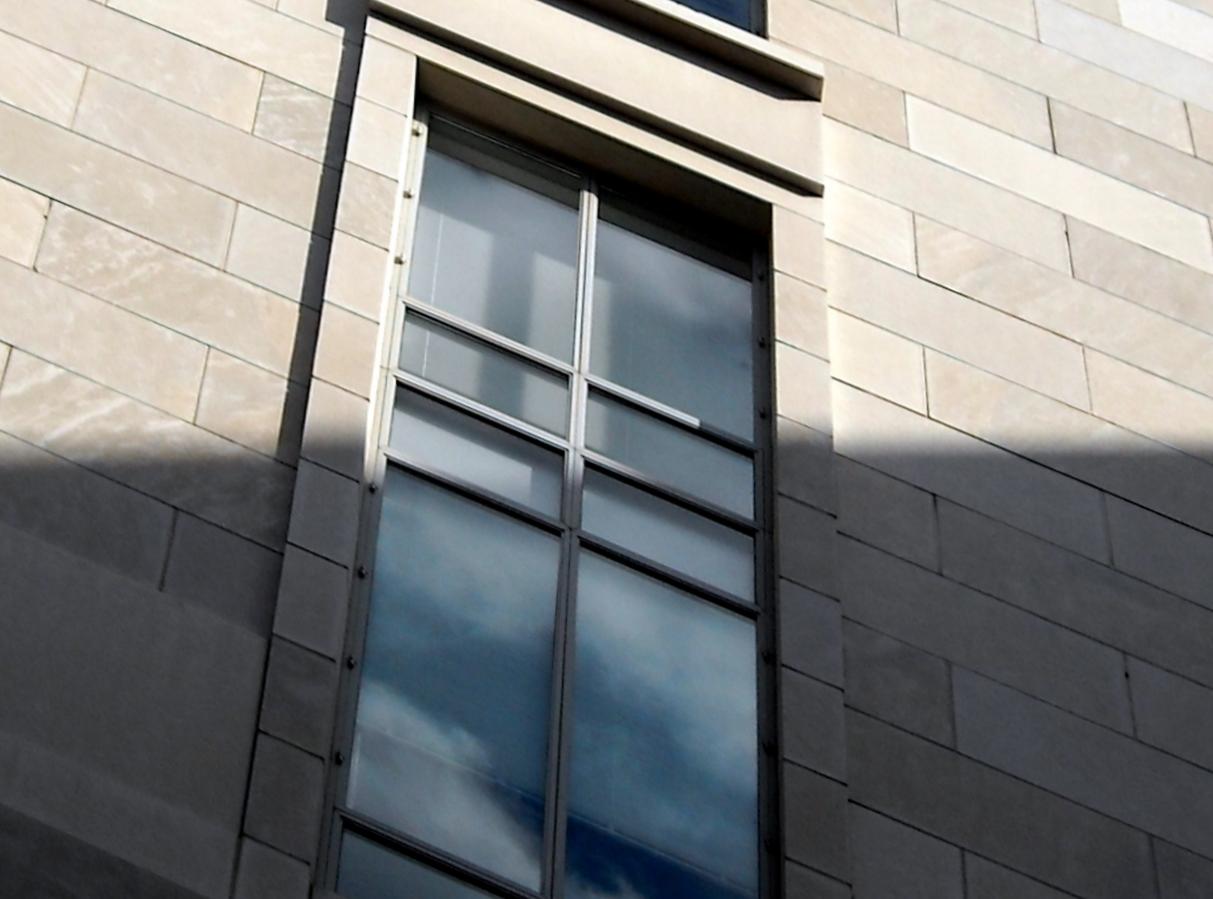}
}
\subfigure{
\includegraphics[width=0.22\textwidth]{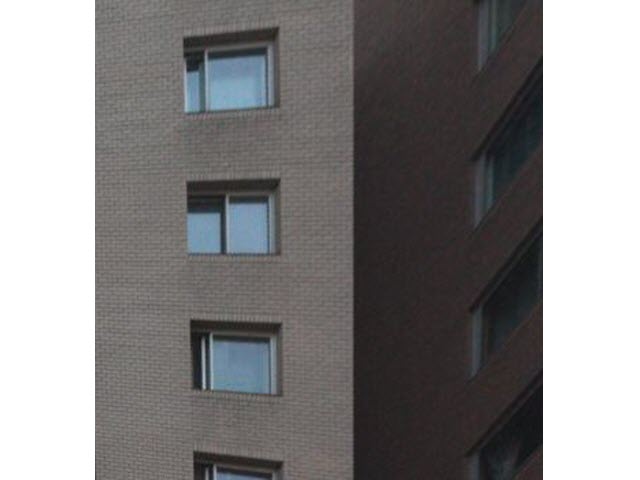}
}
\subfigure{
\includegraphics[width=0.24\textwidth]{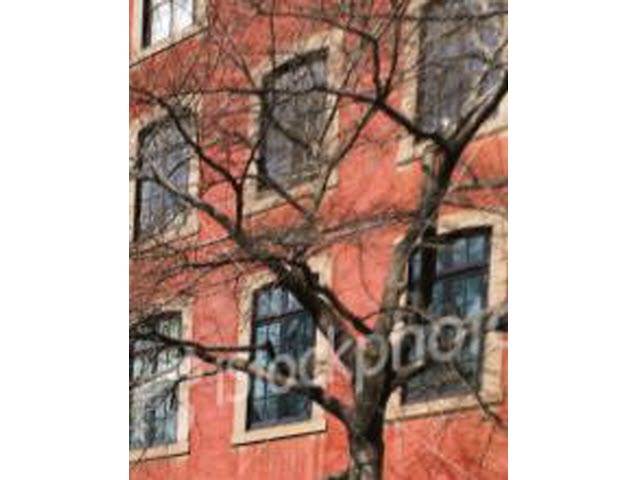}
}
\subfigure{
\includegraphics[width=0.22\textwidth]{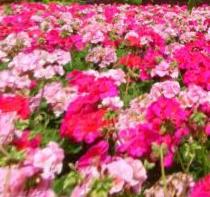}
}
}
\caption{{\bf Failure Cases.} TILT fails to recover the geometry of these images since they deviate from the assumptions under which TILT is designed to work. From left to right: two incompatible dominant low-rank structures, overlapped or adjacent; too much occlusion; random textures.}
\label{fig:failed-cases}
\end{figure*}

\paragraph{Expected failures.}
It should come as no surprise that when the assumptions of TILT are violated, it no longer finds the low-rank structure correctly. Figure \ref{fig:failed-cases} shows the results of TILT on some examples:
\begin{itemize}
\item The first example (Figure \ref{fig:failed-cases}(a)) shows the limitations of the ``low-rank'' assumption on some man-made structures: Two incompatible dominant low-rank structures (the facade and the shadow) are overlapped, which result in an overall high-rank region. TILT actually aligns to the orientation of the strong shadow. In order to make this case work, a simple ``low-rank" promoting objective, like the one in TILT, is no longer sufficient. 
\item The second example (Figure \ref{fig:failed-cases}(b)) shows another limitation of the low-rank assumption. If the chosen window contains two adjacent low-rank regions each of which is distorted differently, the combined region might no longer be low-rank when subject to one global affine or projective transformation. Proper segmentation of the different low-rank regions is needed before TILT can work correctly on each of the low-rank regions; or TILT has to be extended to simultaneously handle multiple domain transformations.
\item Although TILT is designed to be robust to corruptions or occlusions, it is effective only when the amount of corruption is not too large. As shown in Figure \ref{fig:failed-cases}(c), if there is too much occlusion, TILT cannot be expected to succeed, even though human vision is still capable of perceiving the building structures behind the tree. It remains to be seen whether the robustness of TILT can be improved to handle such challenging cases.
\item As mentioned earlier in Section \ref{sec:formulation}, TILT is not designed to work on random textures, such as the one shown in Figure \ref{fig:failed-cases}(d). Although there has been work in the literature showing that it is possible to infer approximate orientation of the flower bed based on statistical property of the random texture, TILT is certainly not designed to handle such cases -- it is effective only for regular symmetric textures, but not for random textures.
\end{itemize}

\section{Potential Modifications and Extensions}
\label{sec:extensions}
The TILT algorithm proposed in this paper is still rather rudimentary. Nevertheless, due to its simplicity, it can be easily modified or extended to handle more complex scenarios in natural images. In this section, we demonstrate this with three possible extensions. The reader should be aware that we do not claim that we have already given the best solution to each problem discussed here. Instead, the goal is merely to show the readers some basic ideas about how to modify TILT. In fact, we believe, each of the problem deserves a much more thorough investigation so that more effective and efficient algorithms could be developed in the future. 

\subsection{Matrix Completion for Boundary Effects}
We note that in Step 3 of Algorithm \ref{alg:outer_loop}, we update the transformation parameters $\tau$, and recompute the transformed image $I\circ \tau$ in Step 1 of the subsequent iteration. While this is conceptually sound, it poses a serious problem in practice. This is because real images always have finite support or size. So, if the window containing the texture of interest is close to the image boundary, then the transformed image window $I \circ \tau$ might not be well-defined at all pixels. The conventional methods to treat this problem is to either assume that the region outside the image has zero pixel values, or to interpolate them from the boundary pixels ensuring some degree of smoothness. The former approach is ill-suited to our problem since it may destroy the low-rank structure of the texture inside the image (hence TILT may fail to converge to the correct solution as shown in Figure \ref{fig:challenging-cases}(a)), while the latter introduces more free parameters to the algorithm, namely the choice of the interpolation function.

\begin{figure*}[ht]
\centering{
\includegraphics[width=0.57\columnwidth]{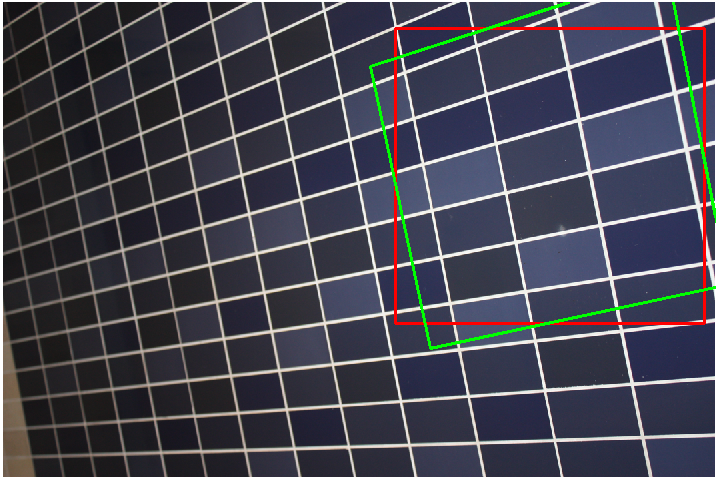}
\includegraphics[width=0.4\columnwidth]{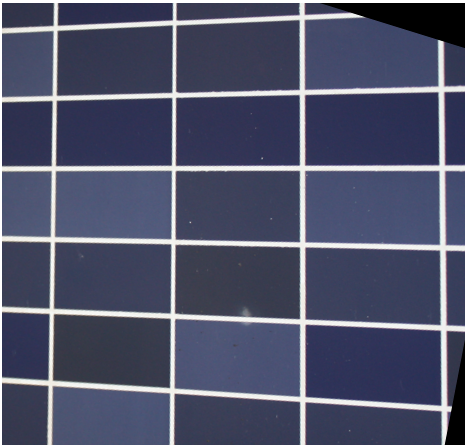}\hspace{3mm}\includegraphics[width=0.57\columnwidth]{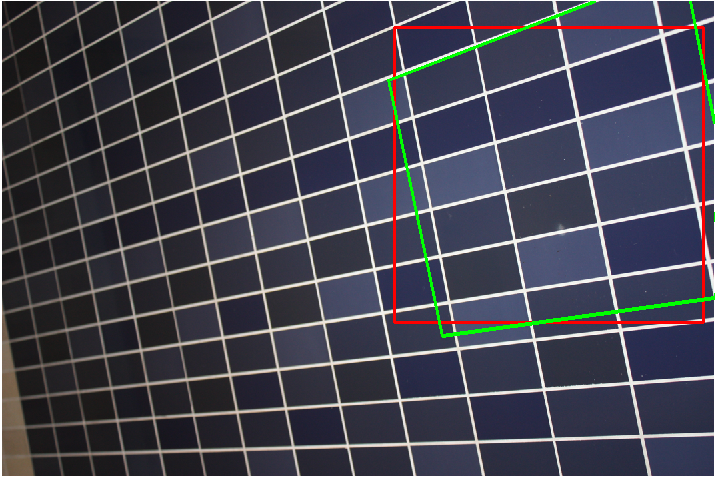}
\includegraphics[width=0.4\columnwidth]{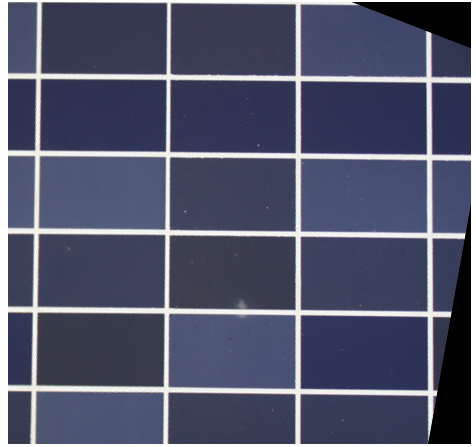}
}
\centering{
\includegraphics[width=0.64\columnwidth]{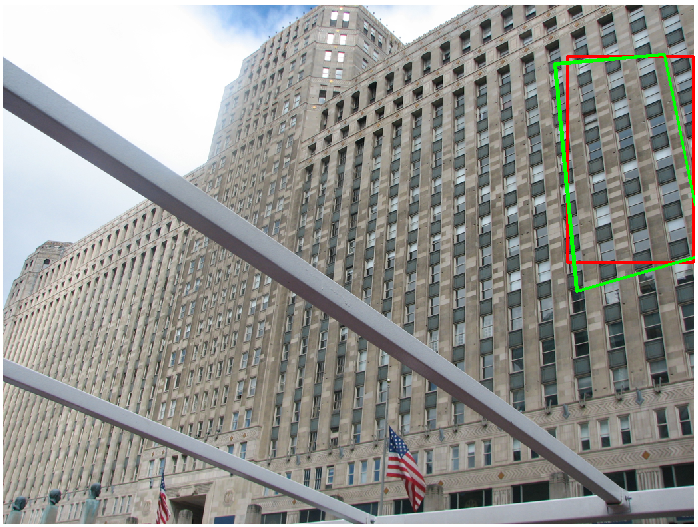}
\includegraphics[width=0.3\columnwidth]{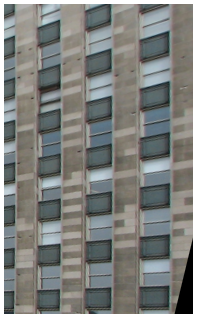}\hspace{3mm}\includegraphics[width=0.64\columnwidth]{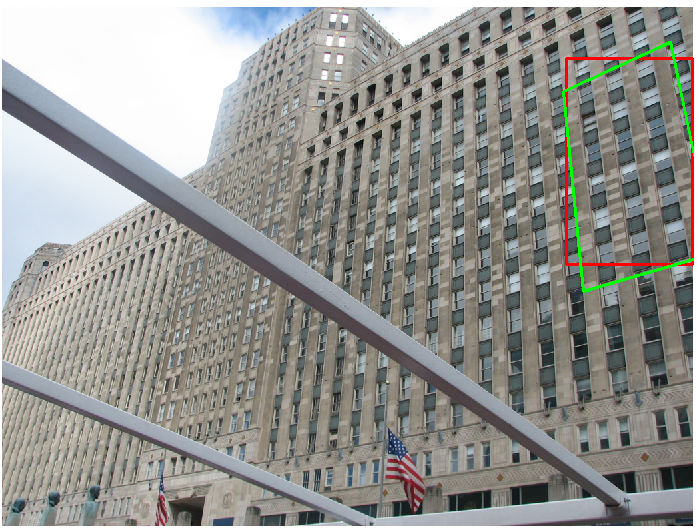}
\includegraphics[width=0.3\columnwidth]{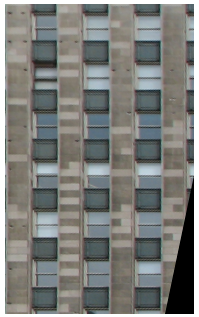}
}
\caption{{\bf TILT with or without Matrix Completion.} Left: Basic TILT without matrix completion -- TILT stops when the region goes over the image boundaries before it converges to the correct transform; Right: TILT with matrix completion -- with the same initialization it converges to the correct transform.}
\label{fig:completion_example}
\end{figure*}

This problem can actually be handled in a more principled manner. We treat the pixels that fall outside the image boundary as missing entries of the low-rank matrix to be recovered. This formulation is in a similar spirit as the low-rank matrix completion problem that has been extensively studied recently \citep{Recht2008-SR,Candes2008,Candes2010-IT}. Let $\Omega$ represent the set of pixels that are located inside the image boundary after transformation. Then, we modify the constraint in the linearized problem \eqref{eqn:convex_linear_tilt} as follows:
\begin{equation}
\pi_{\Omega}(I \circ \tau +\nabla I  \Delta \tau)  = \pi_{\Omega}(I^0 + E),
\end{equation}
where $\pi_{\Omega}(\cdot)$ denotes the projection operator onto the set of entries with support in $\Omega$. Thus, we apply the constraint only on the set of pixels at which the transformed image $I\circ\tau$ is well-defined. Since $\pi_{\Omega}(\cdot)$ is a linear operator, the resulting optimization problem is still a convex program and can be solved by the ALM algorithm outlined in Section \ref{sec:alm}.\footnote{One can even handle small noise in this case, as shown in the work of \cite{Yuan2010-pp}.} Figure \ref{fig:completion_example} shows two examples of how matrix completion could improve the performance of TILT when the chosen window is too close to boundaries of the image and the correct solution needs to converge to outside of the original image.

\subsection{Enforcing Reflective Symmetry} Notice that ``low-rank'' is merely the result of many types of regularities and symmetries. However, a low-rank texture need not necessarily be symmetric. Hence, if we intend to recover a symmetric texture, it might not be sufficient to impose only the ``low-rank'' objective. For instance, although many of the examples seen earlier have reflective symmetry, the axis of symmetry is not necessarily always at the center of the recovered low-rank region. So in order to ensure that the recovered low-rank region has such symmetry, additional constraints need to be imposed on TILT.

Suppose that $I^0 \in \R^{m \times n}$ represents the image of a texture with reflective symmetry. Without any loss of generality, we may assume that the axis of symmetry is horizontal. Then, the reflective symmetry of $I^0$ can be expressed mathematically as
\begin{equation}
I^0(i,j) = I^0(m+1-i,j), \; \forall (i,j) \in \{1: m\} \times \{1: n\}.
\end{equation}
In general, for any type of symmetry for $I^0$, we can find an invertible linear mapping $g:\R^2\rightarrow \R^2$ such that $g(I^0) = I^0$.\footnote{For reflective symmetry, $g$ is its own inverse.} Thus, we may add any desired symmetry as an additional set of constraints to the linearized convex program \eqref{eqn:convex_linear_tilt} in the TILT framework. Since the constraints from symmetry are all linear in $I^0$, we can easily use the ALM algorithm described in Section \ref{sec:alm}, with minor modifications, to solve the new constrained optimization problem.

\begin{figure*}[!ht]
\centering{
\includegraphics[height=0.36\columnwidth]{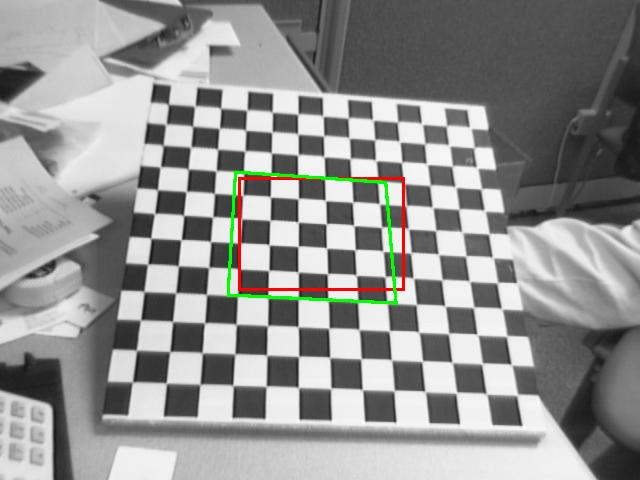}\hspace{2mm}
\includegraphics[height=0.25\columnwidth]{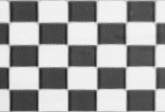}\hspace{2mm}
\includegraphics[height=0.25\columnwidth]{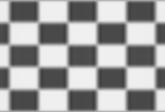}\hspace{2mm}
\includegraphics[height=0.25\columnwidth]{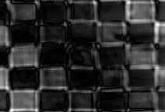}
}\\
\centering{
\includegraphics[height=0.65\columnwidth]{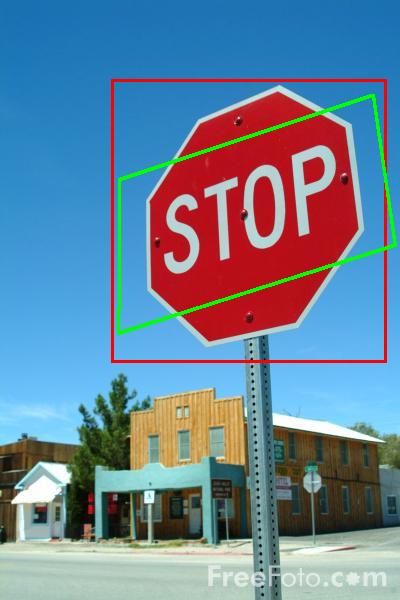}\hspace{2mm}
\includegraphics[height=0.4\columnwidth]{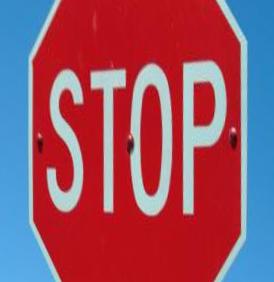}\hspace{2mm}
\includegraphics[height=0.4\columnwidth]{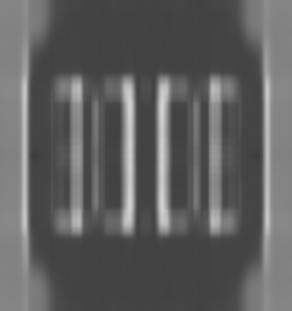}\hspace{2mm}
\includegraphics[height=0.4\columnwidth]{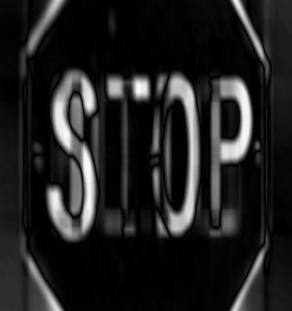}
}
\caption{{\bf Reflective Symmetry Imposed.} Top row: results on a checker-board. From left to right: the original image $I$; the rectified image $I\circ \tau$; the recovered low-rank component $I^0$; and the sparse component $E$. Bottom row: The corresponding results of the stop sign example (in Figure \ref{fig:challenging-cases}) with reflective symmetry enforced.}
\label{fig:symmetry_example}
\end{figure*}

We have implemented a modified version of TILT which enforces the recovered low-rank component $I^0$ to have reflective symmetry in both $x$ and $y$-directions.\footnote{In order to allow the low-rank region to move freely to a symmetric region, we have to remove the constraints on the translation.} Figure \ref{fig:symmetry_example} (top) shows the result of the modified algorithm on a checker-board  with reflective symmetry in the $x$ and $y$-directions enforced: Notice that the converged region is indeed symmetric in both directions. Figure \ref{fig:symmetry_example} bottom shows the new converged results of the same stop sign example in Figure \ref{fig:challenging-cases} with the same initialization. Notice that this is in fact a very challenging case for TILT as the foreground (the sign) is very sparse in the image domain. The recovered low-rank part $A$ is indeed very symmetric and the sparse part $E$ accounts for all sparse deviations from the symmetry (including asymmetry in the letters).

\subsection{TILT for Rotational Symmetry} Many other structural properties may be converted to a low-rank objective. For instance, the image of a rotationally symmetric pattern need not be a low-rank matrix, but it can be converted to one. To deal with rotational symmetry, we will consider circular windows, instead of rectangular ones. Each circular window is uniquely determined by its center and its radius. Clearly, the image region enclosed by such a window is not a matrix. However, it can be converted to one by considering a Frieze-expansion pattern (FEP) of the region \citep{Liu2004-PAMI,Lee2010-PAMI}.

\begin{figure*}
\centerline{
    \subfigure[Input image and initial circle]
    {
        \includegraphics[scale=0.3]{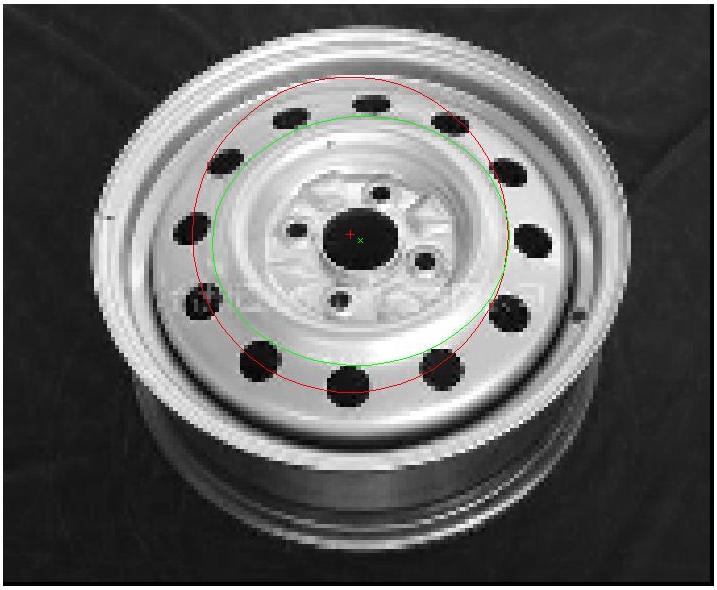}
    }
    \subfigure[Frieze-expansion patterns]{
        \includegraphics[scale=0.7]{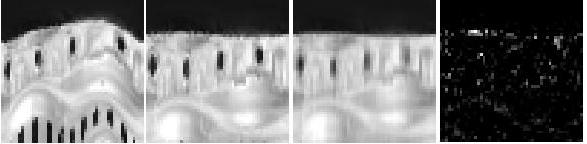}
    }
}
\caption{{\bf Rotational Symmetry with Affine Transform.} (a) Input image with inherent rotational symmetry. The symmetry is not immediately evident due to the deformation caused by the viewpoint. The red window denotes the input and the green window encloses the symmetric pattern converged to by our algorithm. (b) From left to right: the FEP of the input $I$ (red) window which does not exhibit any low-rank structure; the FEP of the output $I\circ \tau$ (green) window recovered by TILT; the corresponding low-rank texture $I^0$; and the sparse error $E$ in the recovered FEP.}
\label{fig:rotation}
\end{figure*}

Suppose that a matrix $I^0 \in \R^{m \times n}$ is the FEP of a circular window in an image with center at the origin and radius $R$. Then, the mapping $\tau$ between an entry $(x_0,y_0)$ in $I^0$ and its corresponding pixel in the image is given by
\begin{equation}
\tau\left(x_0,y_0\right) = \left(\frac{Rx_0}{m}\cos{\Big(\frac{2\pi y_0}{n}\Big)}, \frac{Rx_0}{m}\sin{\Big(\frac{2\pi
y_0}{n}\Big)}\right).
\label{eqn:fep}
\end{equation}
If the center and radius of the circular window are chosen correctly, then the above FEP mapping gives rise to a low-rank matrix. However, in practice, the exact position of the window is not known a priori. In addition, there could be an additional deformation of the pattern due to the viewpoint. Figure \ref{fig:rotation}(a) shows a representative input image. Suppose we model the deformation by an affine transformation. Then, the mapping \eqref{eqn:fep} from the low-rank matrix to the input image can be rewritten as
\begin{equation}
\tau\left(x_0,y_0\right) = H\cdot \left[\frac{Rx_0}{m}\cos{\Big(\frac{2\pi y_0}{n}\Big)}, \frac{Rx_0}{m}\sin{\Big(\frac{2\pi
y_0}{n}\Big)},1\right]^T,
\label{eqn:affine_fep}
\end{equation}
where $H$ represents an affine transformation in homogenous coordinates. We can easily modify TILT to deal with the combined deformation of the FEP and the affine map and the algorithm can simultaneously recover the correct center of symmetry and the affine deformation. We show the results of such an algorithm on one rotationally symmetric pattern in Figure \ref{fig:rotation}.

\section{Conclusions and Future Directions}
In this paper, we introduce a novel framework in which an image window is viewed as a matrix and the rank of the matrix is used as a measure of textural simplicity in the image window. We have introduced a very effective way of extracting precise structure and geometry of low-rank textures from their images using iterative convex optimization techniques. The proposed algorithm works effectively and robustly for a wide range of regular, symmetric patterns and structures in real images, suggesting that the {\em transformed low-rank plus sparse structures model} is important for modeling real images of urban environments and man-made objects. More importantly, the proposed tools are highly complementary to most existing vision techniques that mainly focus on using local features. Instead,  by leveraging efficient high-dimensional optimization techniques, the new tools can process a large image region to extract dominant structural and geometric information more accurately and robustly in a holistic fashion.

The proposed TILT scheme is still quite rudimentary in its formulation and solution. Many aspects of it can still be improved. Also, it can be customized or extended by incorporating additional structural constraints or by considering different  deformation models. Conceptually, there should be little difficulty in generalizing TILT from the linear (affine or projective) transforms to other classes of possibly nonlinear domain deformations. An important open problem is to derive conditions (on the type of signals and the deformation groups) under which this framework is guaranteed to succeed. As being low-rank is only a necessary but not sufficient property for many regular, symmetric patterns and structures, it is worth investigating in the future more pertinent measures or objective functions for recovering such patterns and structures despite geometric deformation. More generally, this work could motivate people to discover new types of (transform-invariant) properties that can be extracted effectively and efficiently from images in a similar holistic fashion, without relying on local features.

The low-rank textures and the associated geometric transformations recovered by TILT can be very useful for many high-level computer vision tasks such as image compression, matching, segmentation, symmetry detection, reconstruction of 3D models of urban environments, and recognition of man-made objects. On the other side of the coin, in this paper we have not fully addressed the issue of detecting the location and scale of candidate low-rank regions so as to better initialize and apply the TILT algorithm. As some of our experiments have suggested, better initialization can significantly improve the performance and applicability of TILT to a broader range of situations. This leaves plenty of room for future investigation on how to improve and augment TILT with other computer vision techniques such as image segmentation and salient region detection, or with other scale-invariant local features such as SIFT.  

\section*{Acknowledgements}
The authors would like to thank Dr. John Wright of Microsoft Research Asia for many stimulating discussions on this topic and for suggesting the catchy acronym ``TILT'' for the proposed method. We would also like to thank Professor Xiaoming Yuan of Hong Kong Baptist University and Professor Emmanuel Cand\`es of Stanford for sharing their insights on the convex optimization algorithms during the preparation of this manuscript. 

\section*{Appendix A: Derivation of Linear Constraints}
\label{ap:constraints}
In Section \ref{sec:implementation}, we have proposed to impose two sets of constraints on the deformation parameters to make the solution well-defined so as to avoid some pathological solutions. Here, we show a detailed derivation or linearization of these constraints. In particular, we present here the derivation for the case when the transformation group is the set of affine transformations. The derivation for the homography case is very similar in case such constraints need to be imposed.

\paragraph{Constraints on Translation \eqref{eqn:cons1}.}
Our first constraint is that the center of the rectangular window is fixed {\it i.e.}, if $x_0 = [x_0(1) \; x_0(2)]^T$ is the initial center of the window and $\tau$ is the optimal transformation, then $\tau(x_0) = x_0$. Since the transformation is affine, we have that $\tau(x) = Ax + b$, where $A = \left[\begin{array}{cc}A_{11} & A_{12} \\ A_{21} & A_{22}\end{array} \right]$ is an invertible matrix and $b \in \R^2$. Suppose we parameterize our transformation vector as 
$$
\tau = \left [
\begin{array}{c}
A_{11} \\ A_{21} \\ A_{12} \\ A_{22} \\ b
\end{array}
\right ],
$$
then in \eqref{eqn:cons1} we have
\begin{equation} 
A_t = \left [
\begin{array}{cccccc}
x_0(1) & 0 & x_0(2) & 0 & 0 & 0 \\
0  & x_0(1) & 0 & x_0(2) & 0 & 0
\end{array}
\right ].
\end{equation}

\paragraph{Constraints on Scale \eqref{eqn:cons2}.} The second constraint ensures that the area covered by the window as well as its aspect ratio does not change drastically. We show how this constraint results in a linear constraint in $\dt$. For a given affine transformation, we note that the size of a rectangle gets scaled by the same amount. Thus, without any loss of generality, we assume that the initial window is a unit square with the points $(0,0)$ and $(1,1)$ forming opposite diagonal vertices. Once again, we represent the affine transformation by $\tau(x) = Ax + b$. Let $S(A,b)$ denote the area of the window after transformation. Since the area of the window is unchanged by translation, we denote the area as $S(A)$. Let $e_1$ and $e_2$ denote two adjacent edges (with the origin as the common vertex) of the initial square. After transformation, these edges can be represented by the vectors $e_1 = (A_{11},A_{21})$ and $e_2 = (A_{12},A_{22})$. Then, the area of the transformed window is given by
\begin{equation}
S(A) = \frac{1}{2} \,\|e_1\| \,\|e_2\| \,\sin\theta,
\end{equation}
where $\cos\theta = \frac{\langle e_1,e_2\rangle}{\|e_1\|\,\|e_2\|}$. The above equation can be simplified to
\begin{equation}
S(A) = \sqrt{\left(A_{11}A_{22} - A_{12}A_{21}\right)^2}.
\end{equation}
Now suppose that the matrix $A$ is perturbed by a small amount $\Delta A$. Since we require that the new area $S(A+\Delta A)$ is close to $S(A)$, we impose the constraint that the first-order term in the Taylor series expansion of $S(A+\Delta A)$ be zero {\it i.e.},
\begin{equation}
\nabla_A \,S(A)\cdot \Delta A = 0.
\label{eqn:area}
\end{equation}

We now consider the second part of the constraint which is to minimize the rate at which the aspect ratio of the window changes. SInce the aspect ratio is unity for the initial window, we essentially require that $\|e_1\| = \|e_2\|$ for the transformed window, using the same notation as above. We define $C(A) = \|e_1\|^2 - \|e_2\|^2$. Then, ideally, we require $C(A+\Delta A)$ to be close to zero. Once again, we impose the constraint that the first-order term in the Taylor series expansion to be zero {\it i.e.},
\begin{equation}
\nabla_A \,C(A)\cdot \Delta A = 0.
\label{eqn:ratio}
\end{equation}

Combining \eqref{eqn:area} and \eqref{eqn:ratio}, and denoting $\tau$ as a vector of all the transformation parameters, it is easy to see that we get a linear constraint of the form $A_s \dt = 0$, as given in \eqref{eqn:cons2}.

\bibliographystyle{spbasic}
\bibliography{TILT_IJCV}

\end{document}